%% file: main.tex
\documentclass{article}

\PassOptionsToPackage{numbers,sort&compress}{natbib}
\usepackage[final]{neurips_2025}

\newcommand{\appendixtitle}{
    \clearpage
    \begin{center}
        \LARGE\bfseries - Appendices -
    \end{center}
    \addcontentsline{toc}{section}{Appendices} 
    \setcounter{section}{0}
    \renewcommand{\thesection}{\Alph{section}}
    \renewcommand{\thesubsection}{\thesection.\arabic{subsection}}
}

\usepackage[utf8]{inputenc} 
\usepackage[T1]{fontenc}    
\usepackage{url}            
\usepackage{booktabs}       
\usepackage{amsfonts}       
\usepackage{nicefrac}       
\usepackage{microtype}      
\usepackage{xcolor}         
\usepackage[pagebackref=true,breaklinks=true,colorlinks=true,bookmarks=false,linkcolor={red!50!black},urlcolor={darkgray!80!black},citecolor={green!50!black}]{hyperref} 

\input{macros}
\input{preamble}

\title{DGH: Dynamic Gaussian Hair }

\author{%
  \textbf{Junying Wang}\textsuperscript{1,2\thanks{Work performed during an internship at Meta Reality Labs Research}}\quad
  \textbf{Yuanlu Xu}\textsuperscript{2}\quad
  \textbf{Edith Tretschk}\textsuperscript{2}\quad
  \textbf{Ziyan Wang}\textsuperscript{2}\quad
  \textbf{Anastasia Ianina}\textsuperscript{2}\\
  \textbf{Aljaz Bozic}\textsuperscript{2}\quad
  \textbf{Ulrich Neumann}\textsuperscript{1}\quad
  \textbf{Tony Tung}\textsuperscript{2}\\\\
  {
    \textsuperscript{1}University of Southern California \quad\quad
    \textsuperscript{2}Meta Reality Labs Research
  }%
}

\begin{document}

\maketitle

\begin{center}
\vspace{-2mm}
\includegraphics[width=\textwidth]{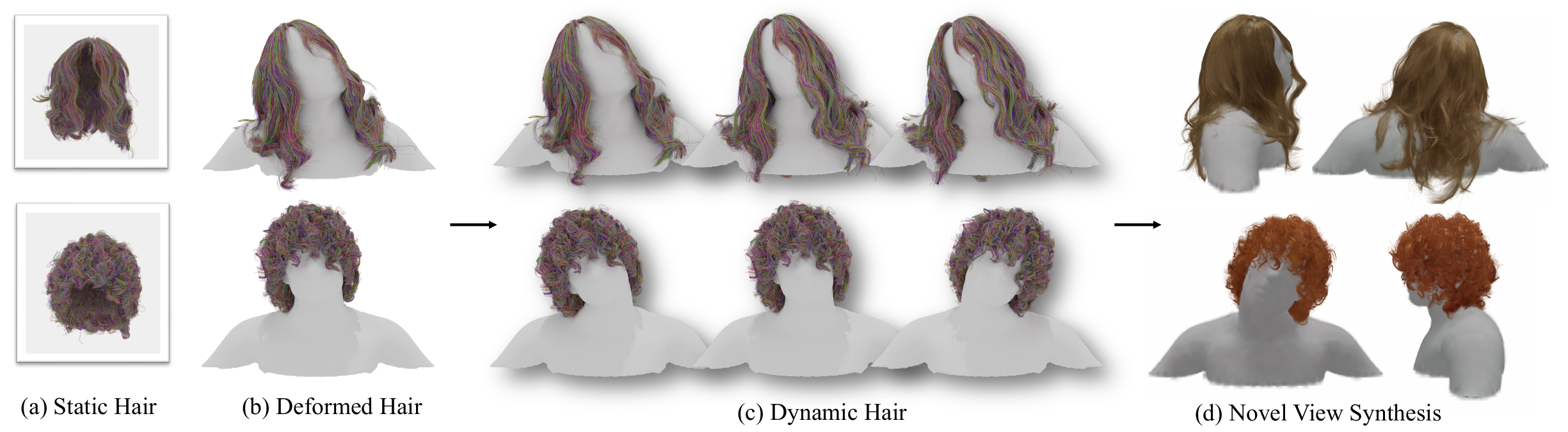}
\vspace{-2mm}
\captionof{figure}{
Dynamic Gaussian Hair (\dgh) is a framework that learns dynamic deformation and photorealistic novel-view synthesis of arbitrary hairstyles driven by head motions, while respecting upper-body collision.
At runtime, given a hairstyle and head motion (a), DGH infers initial hair deformations (b), refines the deformations with dynamics (c), and generates 3D Gaussian Splats to achieve photorealistic novel-view synthesis (d).
}
\label{fig:teaser}
\end{center}

\input{sec/0_abstract}

\input{sec/1_intro}

\input{sec/2_related_work}

\input{sec/3_method}

\input{sec/4_experiment}

\input{sec/5_conclusion}

\clearpage
\bibliographystyle{unsrtnat}
\bibliography{ref}


\newpage
\input{sec/X_suppl}



\end{document}

%% file: macros.tex
\usepackage{xcolor}
\usepackage{enumitem}
\usepackage{xspace}
\usepackage{booktabs}
\usepackage{colortbl}
\usepackage{multirow}
\usepackage{makecell}
\usepackage{lipsum} 
\usepackage{gensymb} 
\usepackage{graphicx}
\usepackage{amssymb}
\usepackage{amsmath}
\usepackage{wrapfig}
\usepackage{float}
\usepackage{multicol}
\usepackage{appendix} 
\usepackage{algorithm}
\usepackage{algorithmic}
\usepackage[table]{xcolor}
\definecolor{YellowGreen}{rgb}{0.75, 0.85, 0.65}

\definecolor{MyDarkRed}{rgb}{0.66, 0.16, 0.16}
\definecolor{MyDarkBlue}{rgb}{0.16, 0.16, 0.66}

\makeatletter
\DeclareRobustCommand\onedot{\futurelet\@let@token\@onedot}
\def\@onedot{\ifx\@let@token.\else.\null\fi\xspace}

\makeatother

\usepackage[labelsep=period]{caption}
\renewcommand{\paragraph}[1]{\vspace{0.05cm}\noindent \textbf{#1 \hspace{0.2em}}}



%% file: preamble.tex
\newcommand{\dgh}{DGH\xspace}

%% file: sec/0_abstract.tex
\begin{abstract}


The creation of photorealistic dynamic hair remains a major challenge in digital human modeling because of the complex motions, occlusions, and light scattering. 
Existing methods often resort to static capture and physics-based models that do not scale as they require manual parameter fine-tuning to handle the diversity of hairstyles and motions, and heavy computation to obtain high-quality appearance.
In this paper, we present Dynamic Gaussian Hair (\dgh), a novel framework that efficiently learns hair dynamics and appearance. We propose:
(1) a coarse-to-fine model that learns temporally coherent hair motion dynamics across diverse hairstyles;
(2) a strand-guided optimization module that learns a dynamic 3D Gaussian representation for hair appearance with support for differentiable rendering, enabling gradient-based learning of view-consistent appearance under motion. 
Unlike prior simulation-based pipelines, our approach is fully data-driven, scales with training data, and generalizes across various hairstyles and head motion sequences. 
Additionally, DGH can be seamlessly integrated into a 3D Gaussian avatar framework, enabling realistic, animatable hair for high-fidelity avatar representation.
DGH achieves promising geometry and appearance results, providing a scalable, data-driven alternative to physics-based simulation and rendering. Our project page: \href{https://junyingw.github.io/paper/dgh}{\texttt{https://junyingw.github.io/paper/dgh}}


\end{abstract}

%% file: sec/1_intro.tex
\section{Introduction}
\label{sec:intro}


Realistic, high-fidelity dynamic hair modeling and rendering are crucial for creating realistic avatars in animation and AR/VR applications. Adding hair dynamics can significantly enhance the realism of animated characters, as hair deformation and high-order motion effects contribute to physical plausibility.
Capturing and modeling these effects, however, poses substantial challenges due to the complexity of hair motion and the intricate interactions among hair strands and head/shoulder. 
Physics-based hair simulation methods produce high-quality hair dynamics but lack generalization as they require manual parameter fine-tuning per hairstyle (e.g., ponytail, curly hair) to reproduce realistic deformations (e.g., hair stiffness, density, damping). They also need to model interactions with the environment, such as collisions with shoulders, which are computationally expensive. The models are non-differentiable, and thus unsuitable for scalability with data-driven based solutions. 
Robust implementation of complex deformations is nontrivial, and unexpected behaviors usually break the experience. Most real-time applications model only simple or quasi-static deformations.

\begin{figure*}[ptb]
\centering
\includegraphics[width=1\textwidth]{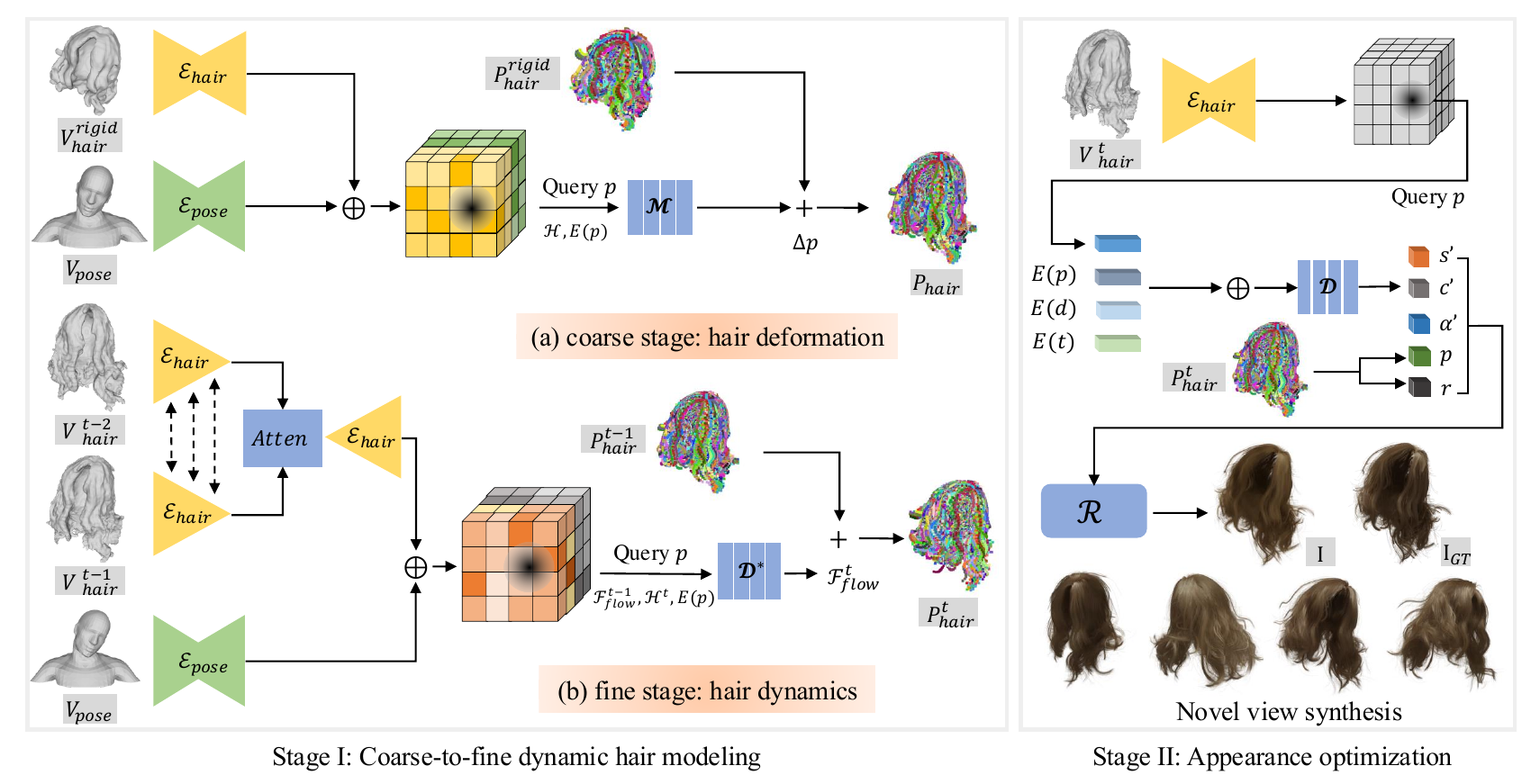}
\vspace{-3mm}
\caption{\textbf{Framework Overview.} 
\dgh learns hair deformation dynamics and photorealistic appearance. \textbf{Stage I: Coarse-to-Fine Dynamic Hair Modeling.} The input hair model and the upper body are transformed into a canonical hair volume \( V^{\text{rigid}}_{\text{hair}} \) and a pose volume \( V_{\text{pose}} \), respectively. Then, a coarse-to-fine strategy deforms the hair model. At the coarse stage, points \( \mathbf{p}_i \) are sampled from the rigidly transformed hair, and the interpolated features from \( \mathcal{E}_{\text{pose}} \), \( \mathcal{E}_{\text{hair}} \), head pose \( \mathcal{H} \), and positional encoding \( E(p) \) are concatenated and fed into an MLP \( \mathcal{M} \) to predict displacements \( \Delta \mathbf{p} \), producing deformed hair points \({P}_{\text{hair}} \). The fine stage refines hair deformation with dynamics by estimating flow \( \mathcal{F}_{\text{flow}}^t \) through cross-attention between volumetric features from previous frames \( V_{\text{hair}}^{t{-}2} \) and \( V_{\text{hair}}^{t{-}1} \), ensuring smooth temporal transitions.
\textbf{Stage II: Appearance Optimization.} We train an MLP \( \mathcal{D} \) to predict color \( c' \), scale \( s' \), and opacity \( \alpha' \) of 3D Gaussian Splats from features of the deformed hair. Differentiable rasterization leverages the appearance model to synthesize high-quality renderings that adapt to hair movement and occlusion dynamics.
}
\label{fig:overview}
\vspace{-3mm}
\end{figure*}
The computational cost to create photorealistic digital hair is very high. Realistic rendering requires path tracing of accurate hair modeling, which is translucent, contains occlusions, and accounts for light scattering effects~\cite{marschner2003light}.
Hair occlusion patterns change while in motion, posing significant challenges for dynamic hair rendering.
The creation of a high-quality synthetic hair dataset at scale typically requires a rendering farm with multiple GPUs.
Real-time hair rendering engines (Unreal, Unity) compromise on quality (e.g., approximate lighting) while still requiring a desktop GPU.

Methods relying on Neural Radiance Fields (NeRF)~\cite{mildenhall2021nerf,park2021nerfies} or 3D Gaussian Splatting (3DGS)~\cite{qian2023gaussianavatars, xu2023gaussianheadavatar, rivero2024rig3dgs, dhamo2024headgas} have enabled effective and high-quality view synthesis, baking complex hair rendering effects under constrained lighting environments. Although they enable photorealistic avatar re-animation, the hairstyle typically undergoes only rigid transformations without non-rigid or dynamic effects, especially for longer hair, due to weak control signals like head poses of a single timestep.
Similarly, GaussianHair~\cite{luo2024gaussianhair} and Gaussian Haircut~\cite{zakharov2024human} reconstruct hair geometry (from static scenes) using strand-aligned 3D Gaussian representations that can be animated with a physics-based simulation engine, which in turn is usually computationally intensive and requires tedious parameter tuning.

To address these limitations, we introduce \emph{Dynamic Gaussian Hair}, a novel learning-based framework that learns hair deformation with dynamics given canonical hairstyles and head motions. By attaching 3D Gaussians to hair segments and leveraging the appearance optimization, \dgh enables high-quality dynamic hair novel view synthesis at a fraction of the computational cost of a high-end rendering system.
Unlike physics-based simulators, which rely on explicit meshes and require additional rendering and conversion steps, our \dgh framework only needs head rotation and a dense point cloud/pre-trained GS avatar. By converting static hair and upper-body into volume, we enable mesh-free deformation prediction, making our model more compatible with Gaussian-based avatars and easier to integrate into learning-based re-animation pipelines without additional rigging or simulation overhead.

Due to the lack of real captures of dynamic hair deformations with accurate strand tracking, we create a new synthetic dynamic hair dataset from scratch.
Hairstyles with strands are modeled from industry experts, and animated with a physics-based simulation engine. Mutiple-view images are generated with a render farm.
The dataset includes frame-by-frame hair geometry deformation for various hair styles, along with corresponding upper body geometry and head motions. The 3D models are rendered using fine-tuned hair shaders, resulting in photorealistic videos (see Appendices). 

As shown in Fig.~\ref{fig:overview}, our dynamic Gaussian hair framework learns hair deformation dynamics and photorealistic appearance. In the first stage, we introduce a coarse-to-fine framework for \emph{dynamic hair modeling} using a feed-forward network, which, unlike traditional physics-based simulation pipelines, is fully data-driven, differentiable, and free of manual parameter tuning. Specifically, we learn a volumetric implicit deformation model in canonical space that maps static hair geometry to deformed hair, representing hair as a dense point cloud to support diverse hairstyles (e.g., long, curly, ponytail) and accommodate point, Gaussian, or mesh-based representations. 
We encode the head and upper body as volumetric features. During training, a random subset of static hair points is supervised using synthetic ground-truth displacements. The coarse stage is time-independent and serves as a physically plausible hair deformation initialization, leveraging upper-body features and an SDF constraint to prevent hair-body penetration. In the fine stage, we introduce time-dependent dynamics by predicting flow vectors and applying latent-space cross-attention over previous frames to capture temporal consistency and high-frequency hair motion effects such as inertia, oscillation, and damping.
In the second stage, we optimize the \emph{dynamic hair appearance} for novel view synthesis, enabling photorealistic rendering under complex motions and viewpoints. Each hair strand is represented as a sequence of stretched cylindrical Gaussian primitives as shown in Fig.~\ref{fig:curvature}, and we introduce a lightweight non-linear model that refines appearance using strand-level tangent information. Our data-driven framework robustly handles arbitrary hairstyle deformation under varying head motions and occlusions, achieving high-fidelity, view-consistent rendering. Our results are best viewed in our supplemental video. 

We summarize our contributions as:
(1) \textbf{Learning-based volumetric hair deformation:} We introduce a pose-driven, volumetric implicit deformation model that learns to map static hair to dynamic motion across diverse hairstyles and strand densities. Unlike physics-based methods, our approach is fully data-driven, requires no manual parameter tuning, and generalizes to novel head poses.
(2) \textbf{Coarse-to-fine hair deformation dynamics:} We propose a coarse-to-fine learning framework that first predicts pose-dependent deformations and then refines temporal dynamics via flow-based residual learning. This approach ensures stability and generalizes well to unseen head motions.
(3) \textbf{Differentiable dynamic hair appearance optimization:} We represent hair as cylindrical Gaussian primitives and optimize their dynamic appearance with a lightweight, strand-guided network. This differentiable formulation enables photorealistic, view-consistent rendering under motion and occlusion, and integrates seamlessly with neural avatar systems.
We will release our synthetic dynamic-hair dataset to accelerate research on dynamic hair modeling.

%% file: sec/2_related_work.tex
\section{Related Work}
\label{sec:related_work}

Creating realistic avatars includes several challenging steps. One of them is hair capture and animation. Modeling hair appearance and geometry has been studied in the static and dynamic setting. 

\paragraph{Static Hair Modeling}
Handling diverse hairstyles is challenging. Some methods focus on specific cases like braided~\cite{hu2014capturing}, curly~\cite{shao2015modeling}, or generative approaches~\cite{zhou2023groomgen}, while others aim to generalize across static and dynamic settings. They rely on single-view~\cite{chai2015high,chai2016autohair,hu2015single} or multi-view inputs with orientation maps~\cite{paris2004capture, wei2005modeling, hu2014robust, zhou2024groomcap}, sometimes using simulated strands~\cite{luo2012multi, hu2014robust} or geometric heuristics.
NeuralStrands~\cite{rosu2022neural} constrains 3D surface orientation and uses point-based differentiable rendering~\cite{ruckert2022adop, yifan2019differentiable}. NeuralHDHair~\cite{wu2022neuralhdhair} and earlier work~\cite{yang2019dynamic} leverage 2D supervision with 3D spatial cues.
Some methods use real wigs~\cite{shen2023ct2hair}, line-based 3D reconstruction~\cite{nam2019strand}, or OLAT images~\cite{sun2021human} for static hair modeling. Others rely on 2D observations to learn hair growth fields~\cite{kuang2022deepmvshair,wu2022neuralhdhair}, or use optimization and differentiable rendering to extract strands~\cite{takimoto2024dr}.
Beyond image-based approaches, volumetric and neural representations have been explored~\cite{mildenhall2021nerf,zielonka2023instant,sklyarova2023neural}. 
Neural Radiance Fields (NeRF) enable high-fidelity rendering for static scenes~\cite{mildenhall2021nerf,park2021nerfies}, but suffer from slow rendering speeds. Recent representations like Mixture of Volumetric Primitives (MVP)~\cite{lombardi21mvp} and 3D Gaussian Splatting (3DGS)~\cite{kerbl20233d} improve quality and real-time performance, yet dynamic hair reconstruction remains challenging.
INSTA~\cite{zielonka2023instant} models dynamic neural radiance fields with neural graphics primitives around a face model, focusing on head geometry and efficiency. Sklyarova et al.~\cite{sklyarova2023neural} reconstruct hair using implicit volumes followed by strand-level refinement guided by hairstyle priors.
Volumetric hair representations~\cite{wang2021learning, wang2024local} are able to encode 3D global spatial information for static hair modeling. However, all existing work focuses on static hair modeling, leaving the challenge of reanimating static hair for animatable avatars in AR/VR applications an unresolved problem.

\paragraph{Dynamic Hair Modeling}
The motion patterns of hair are hard to emulate. 
One option is hair simulation to advance between adjacent frames with temporal consistency~\cite{zhang2012simulation}. 
In~\cite{xu2014dynamic}, per-frame reconstructions of hair strands are aligned with motion paths of hair strands from spatio-temporal slices of a video volume. 
Grid search over different simulation parameters can be used to determine the set of parameters that matches the visual observations best~\cite{hu2017simulation}. 
Current physics-based and neural hair simulators~\cite{lyu2020real, hsu2024real, daviet2023interactive} enable real-time simulation of hundreds of thousands of strands. However, achieving this on AR/VR devices remains impractical due to their limited GPU memory and processing power. Recent work Quaffure~\cite{stuyck2024quaffure} has only handled quasi-static simulation and does not integrate a tractable rendering solution.
Dynamic hair can also be modeled with neural volumetric approaches. HVH~\cite{wang2022hvh} learns a volumetric representation for dynamic capturing, while NeuWigs~\cite{wang2023neuwigs} accounts for head movements and gravity to produce realistic animations across various hairstyles. 

\paragraph{3D Gaussian Head Avatars}
Modeling realistic 3D head avatars is crucial for immersive VR/AR applications. While 3D Gaussian Splatting (3DGS)~\cite{kerbl20233d} has enabled high-fidelity head avatars, dynamic hair modeling remains a significant challenge, as realistic hair must exhibit natural motion, deformation, and interactions.
%
3DGS-based head avatars~\cite{xu2023gaussianheadavatar, qian2024gaussianavatars, rivero2024rig3dgs, kirschstein2025avat3r, teotia2024gaussianheads} leverage learning-based deformation fields to animate facial expressions and head movements. However, with these models the hair remains a static transformation attached to the head, failing to exhibit natural flow or secondary motion.
Several works have explored hair modeling within the Gaussian framework. GaussianHair~\cite{luo2024gaussianhair} captures strand-level structure using cylindrical Gaussians. Gaussian HairCut~\cite{zakharov2024human} combines strand priors with 3DGS for photorealistic rendering, yet both do not model the dynamic hair appearance. 
To enable realistic hair dynamics in reanimation, we propose a dynamic Gaussian hair representation that models arbitrary hairstyle deformation driven by head motion, supporting dynamic hair novel view synthesis.

%% file: sec/3_method.tex
\section{Method}
\label{sec:method}

In this section, we present our method for Dynamic Gaussian Hair modeling, as illustrated in Fig.~\ref{fig:overview}. Our framework consists of two main stages that model hair dynamics and hair appearance. 
In the first stage, we learn pose-dependent hair deformation via a volumetric implicit function conditioned on canonical static hair, head motion, and upper body mesh. This time-independent model supports arbitrary head poses and strand counts, represented as dense point clouds. We further refine temporal hair dynamics by predicting 3D flow vectors and applying latent-space cross-attention across adjacent frames to ensure temporal consistency.
In the second stage, we optimize time-varying hair appearance to handle motion-induced occlusions, enabling photorealistic novel view synthesis.
During inference, we predict hair deformations in a recurrent manner based on the previous timesteps' deformations. Our appearance model then performs novel view synthesis, generating accurate and realistic hair appearance across frames.
\subsection{Coarse-to-Fine Dynamic Hair Modeling}
Modeling realistic hair dynamics is challenging due to complex motion and temporal variation. We address this with a coarse-to-fine learning framework, and show the following hypothesis:

\paragraph{Hypothesis}
Differentiable rendering of dynamic, photorealistic hair requires accurate and temporally consistent hair tracking. Instead of relying on explicit physics-based simulation, we decompose the simulation task into two stages: a coarse stage that learns pose-driven, time-independent deformation for stable initialization, and a fine stage that refines high-frequency dynamics through 3D flow prediction and temporal cross-attention. We validate this hypothesis through the design of Alg.~\ref{alg:coarse-to-fine}.

\begin{algorithm}[htb]
\caption{\textbf{Coarse-to-Fine Dynamic Hair Modeling}}
\label{alg:coarse-to-fine}
\textbf{Input:} Canonical hair $P_\text{hair}^\text{can}$, proxy mesh (head and shoulders), head poses $\{ \mathcal{H}^{t-2}, \mathcal{H}^{t-1}, \mathcal{H}^t \}$ \\
\textbf{Output:} Dynamic hair $P_\text{hair}^t$ at time $t$
\begin{algorithmic}[1]
\STATE \textbf{Coarse Stage (Time-independent, pose-driven deformation)}
\STATE Transform canonical hair: $P_\text{hair}^\text{rigid} \gets \mathcal{T}_\text{rigid}(P_\text{hair}^\text{can}, \mathcal{H})$
\STATE Voxelize rigid hair and proxy mesh: $V_\text{hair}^\text{rigid} \gets \text{SDF}(P_\text{hair}^\text{rigid}),\; V_\text{pose} \gets \text{SDF}(\text{proxy mesh})$
\STATE Encode features: $\mathcal{V} \gets \text{Concat}(\mathcal{E}_\text{pose}(V_\text{pose}), \mathcal{E}_\text{hair}(V_\text{hair}^\text{rigid}))$
\FOR{$\mathbf{p}_i \in P_\text{hair}^\text{rigid}$}
  \STATE Sample feature: $\mathbf{v}_i \gets \text{Interp}(\mathcal{V}, \mathbf{p}_i)$
  \STATE Predict displacement: $\Delta \mathbf{p}_i \gets \mathcal{M}(\mathbf{v}_i, E(\mathbf{p}_i), \mathcal{H})$
  \STATE Update position: $\mathbf{p}_i \gets \mathbf{p}_i + \Delta \mathbf{p}_i$
\ENDFOR
\STATE Set $P_\text{hair} \gets \{ \mathbf{p}_i \}$
\vspace{3mm}
\STATE \textbf{Fine Stage (Temporal flow refinement)}
\STATE Voxelize past hair states: $V_\text{hair}^{t-2}, V_\text{hair}^{t-1} \gets \text{SDF}(P_\text{hair}^{t-2}), \text{SDF}(P_\text{hair}^{t-1})$
\STATE Encode volumes: $\mathcal{V}_{t-2} \gets \mathcal{E}_\text{hair}(V_\text{hair}^{t-2}),\; \mathcal{V}_{t-1} \gets \mathcal{E}_\text{hair}(V_\text{hair}^{t-1})$
\STATE Cross-attend features: $\mathcal{V}_\text{flow} \gets \text{CrossAttn}(\mathcal{V}_{t-1}, \mathcal{V}_{t-2})$
\FOR{$\mathbf{p}_i \in P_\text{hair}^{t-1}$}
  \STATE Sample volume feature: $\mathbf{f}_i \gets \text{Interp}(\mathcal{V}_\text{flow}, \mathbf{p}_i)$
  \STATE Fuse with pose and prior flow: $\hat{\mathbf{f}}_i \gets \text{Concat}(\mathbf{f}_i, \mathcal{H}^{t}, \mathcal{F}_\text{flow}^{t-1}(\mathbf{p}_i), E(\mathbf{p}_i))$
  \STATE Predict flow vector: $\Delta \mathbf{p}_i^t \gets \mathcal{D}^*(\hat{\mathbf{f}}_i)$
  \STATE Apply flow: $\mathbf{p}_i^t \gets \mathbf{p}_i + \Delta \mathbf{p}_i^t$
\ENDFOR
\STATE Update outputs: $P_\text{hair}^t \gets \{ \mathbf{p}_i^t \},\; \mathcal{F}_\text{flow}^t \gets \{ \Delta \mathbf{p}_i^t \}$
\STATE \textbf{return} $P_\text{hair}^t$
\end{algorithmic}
\end{algorithm}
\textbf{Coarse Stage.}  
We first learn pose-dependent hair deformations from a canonical point cloud \(P_\text{hair}^\text{can}\). The canonical hair is rigid transformed using a head pose \(\mathcal{H}\) to produce \(P_\text{hair}^\text{rigid}\), which is voxelized into an SDF volume \(V_\text{hair}^\text{rigid}\). A proxy mesh of the posed head and shoulders is similarly voxelized into \(V_\text{pose}\). These volumes are encoded via 3D CNNs \(\mathcal{E}_\text{hair}\) and \(\mathcal{E}_\text{pose}\), and concatenated to form the latent feature grid \(\mathcal{V}\).
For each point \(\mathbf{p}_i \in P_\text{hair}^\text{rigid}\), we interpolate its feature \(\mathbf{v}_i\) from \(\mathcal{V}\), and input it along with positional encoding \(E(\mathbf{p}_i)\) and pose \(\mathcal{H}\) into an MLP \(\mathcal{M}\) to predict the displacement \(\Delta \mathbf{p}_i\). The updated position \(\mathbf{p}_i\) is computed as \(\mathbf{p}_i + \Delta \mathbf{p}_i\).
We train using a random subset of hair points, with a total loss:
\begin{equation}
\mathcal{L}_{\text{total}} = \lambda_\text{p} \mathcal{L}_{\text{point}} + \lambda_\text{SDF} \mathcal{L}_{\text{SDF}},
\label{eq:loss_fine}
\end{equation}
where \(\mathcal{L}_{\text{point}}\) is the MSE between predicted and ground-truth displacements, and \(\mathcal{L}_{\text{SDF}}\) penalizes collisions with the body mesh using a hair-to-body Signed Distance Fields~\cite{curless1996volumetric}. This stage yields physically plausible, pose-driven deformation that serves as initialization for dynamic refinement. Please see the Appendices for details on the 3D CNN architecture. 

\textbf{Fine Stage.}  
The fine stage refines dynamic hair motion by predicting temporally consistent 3D flow vectors. We first voxelize the past hair states \(P_\text{hair}^{t-2}\) and \(P_\text{hair}^{t-1}\) into SDF volumes \(V_\text{hair}^{t-2}\) and \(V_\text{hair}^{t-1}\), which are encoded via \(\mathcal{E}_\text{hair}\) into latent feature grids. A cross-attention module aggregates these into a fused volume \(\mathcal{V}_\text{flow}\).
For each point \(\mathbf{p}_i \in P_\text{hair}^{t-1}\), we sample its feature \(\mathbf{f}_i\) from \(\mathcal{V}_\text{flow}\), then concatenate it with the current pose \(\mathcal{H}^t\) and previous flow \(\mathcal{F}_\text{flow}^{t-1}(\mathbf{p}_i)\) to form \(\hat{\mathbf{f}}_i\). These features are passed to a refinement network \(\mathcal{D}^*\) to predict the per-point flow vector \(\Delta \mathbf{p}_i^t\), which is applied to update the hair position \(\mathbf{p}_i^t\).
We supervise the predicted flow \(\mathcal{F}_\text{flow}^t\) using an MSE loss against ground-truth flow \(\mathcal{F}_\text{GT}^t\) from synthetic data, with a total loss:
\begin{equation}\small
\mathcal{L} = \mathcal{L}_{\text{flow}} = \text{MSE}(\mathcal{F}^{t}_{\text{flow}}, \mathcal{F}_\text{GT}^t)
\end{equation}
This stage enables temporally consistent hair animation driven by data rather than explicit simulation.

\subsection{Dynamic Hair Appearance Optimization}

To achieve realistic rendering quality while ensuring speed and optimizability, we rely on 3DGS~\cite{kerbl20233d} to represent hair appearance. 3DGS assigns a learnable position \(\mathbf{p}\), spatial scale $\mathbf{S}$ and rotation $\mathbf{R}$ (combined into covariance matrix \(\Sigma\)), opacity $o$, and color $\mathbf{c}$ to each Gaussian $G$:
\begin{equation}\small
\begin{split}
G(\mathbf{x}) &= e^{-\frac{1}{2} (\mathbf{x}-\mathbf{p})^T \Sigma^{-1} (\mathbf{x}-\mathbf{p})}, \quad \Sigma = \mathbf{R} \mathbf{S} \mathbf{S}^T \mathbf{R}^T, \\
&\quad \Sigma' = \mathbf{J} \mathbf{W} \Sigma \mathbf{W}^T \mathbf{J}^T.
\end{split}
\end{equation}
To render novel views, Gaussians are projected onto the image plane using a differentiable Surface Splatting method~\cite{zwicker2001surface}. Given the viewing transform matrix \( \mathbf{W} \) and the Jacobian \( \mathbf{J} \) of the affine approximation, the covariance matrix \( \Sigma' \) of a 3D Gaussian's corresponding 2D Gaussian is defined as above. 
To determine a pixel's color $\mathbf{c}$, \(N {-} 1\) 2D Gaussians are sorted by depth and then composed with \(\alpha\)-blending:
\begin{equation}\small
\mathbf{c} = \sum_{i=1}^{N} \mathbf{c}_i \alpha_i \prod_{j=1}^{i-1} (1 - \alpha_j),
\end{equation}
where \( \mathbf{c}_i \) is the color of each Gaussian, and \( \alpha_i \) is given by multiplying the opacity with the value of the 2D Gaussian with covariance \( \Sigma' \) at the pixel location.

\begin{wrapfigure}{r}{0.45\linewidth}
  \centering
  \vspace{-3mm}
  \includegraphics[width=\linewidth]{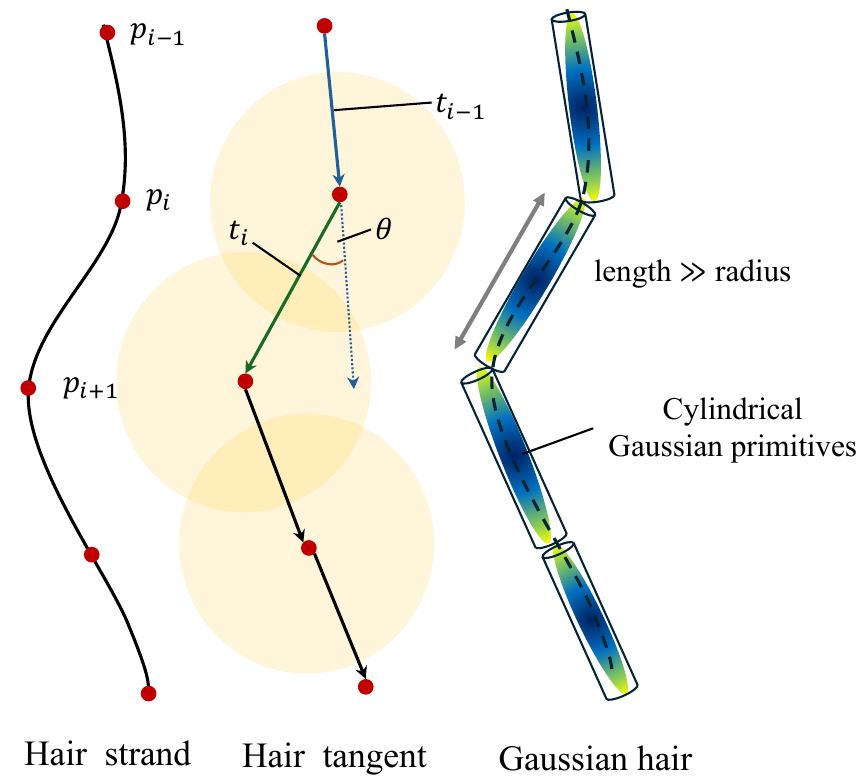}
  \vspace{-4mm}
    \caption{\textbf{Hair Representation.} We show different hair representations with tangent vectors and curvature. For Gaussian hair, we attach cylindrical Gaussian primitive~\cite{luo2024gaussianhair} to each segment with a length much greater than its radius.}
  \label{fig:curvature}
\end{wrapfigure}

Inspired by~\cite{luo2024gaussianhair, zakharov2024human}, we represent each hair strand as a sequence of connected cylindrical Gaussians (Fig.~\ref{fig:curvature}) for dynamic hair rendering, eliminating Gaussian densification~\cite{kerbl20233d}, and fixing the number of primitives across frames. 
Because our dynamic hair model performs forward warping, we can transform each Gaussian from the canonical model to any frame $t$, which makes it trivial to directly propagate the Gaussian color \(\mathbf{c}\) and scale \(\mathbf{s}\) across time. 
The rendering is performed using a Differentiable Tile Rasterizer~\cite{kerbl20233d} \( \mathcal{R} \), which yields image $\mathbf{I}$.

However, Fig.~\ref{fig:compare_app} shows that naively propagating the colors of the canonical frame to the dynamic sequence can lead to missing appearance details in dynamic frames due to complex hair self-occlusion and the interaction of light with hair in motion. 
To address this, we propose a lightweight non-linear model \( \mathcal{D} \) that adjusts Gaussian parameters according to the hair dynamics. 
We train our model using multi-view video sequences under constant lighting. Hair scattering is complex and anisotropic due to strand interactions, as described by the hair BSDF~\cite{marschner2003light, d2011energy}, which highlights the importance of strand tangents in appearance. 
To capture this, we incorporate the hair tangent vector \(\mathbf{t}\) as additional geometric input for optimizing Gaussian parameters.
Specifically, we use an MLP \(\mathcal{D}\) that takes as input the per-point feature sampled from the encoded hair volume \(\mathcal{E}_{\text{hair}}(V_\text{hair}^{t})\) at Gaussian mean \(\mathbf{p}\), along with positional encodings $E$ for $\mathbf{p}$, $\mathbf{t}$, $\mathbf{d}$, and $\mathbf{d}$ denotes the view direction.

These enable \( \mathcal{D} \) to express anisotropic effects. It outputs modified color $\mathbf{c}'$ in the form of spherical harmonic (SH) coefficients, scales $\mathbf{s}'$ and opacity $\mathbf{\alpha}'$ :
\begin{equation}\small
\mathbf{c}', \mathbf{s}', \mathbf{\alpha}'= \mathcal{D} \left( \mathcal{E}_{\text{hair}}(V^t_{\text{hair}}; \mathbf{p}), E(\mathbf{p}), E(\mathbf{t}), E(\mathbf{d}) \right).
\label{eq:appearance}
\end{equation}
As present in Fig.~\ref{fig:overview}, the predicted appearance is then rendered through differentiable rasterization \( \mathcal{R} \) to produce the final image \( \mathbf{I} \). Fig.~\ref{fig:compare_app} shows that our model enhances the hair appearance, achieving realistic renderings under consistent lighting. 
For training, we use common $L_1$, SSIM and LPIPS~\cite{zhang2018perceptual} reconstruction losses:
\begin{equation}\small
L = \lambda_\text{rgb} L_\text{rgb} + \lambda_\text{ssim} L_\text{ssim} + \lambda_\text{lpips} L_\text{lpips}.
\label{eq:loss_render}
\end{equation}

\paragraph{Hypothesis} 
Modeling hair strands with a fixed number of discrete Gaussian primitives leads to visual discontinuities in regions of high curvature. We hypothesize that local curvature is an effective cue for adaptively blending neighboring Gaussians to improve visual continuity.

\noindent\textbf{Curvature-based Gaussian blending.} Unlike continuous surfaces, Gaussian hair treats each segment as an independent shading unit, each with its own tangent vector $\mathbf{t}_{i}$. In high-curvature regions, large angular differences between adjacent tangents ($\mathbf{t}_{i}$, $\mathbf{t}_{i+1}$) can cause shading discontinuities. Although increasing segment density can alleviate this issue, hair tracking typically uses a fixed number of segments per strand. Here we present a simplified hair diffuse shading formulation:
\begin{equation}
\begin{aligned}
I_i &\propto \max(0, \mathbf{t}_i \cdot \mathbf{l}), \\
|I_i - I_{i+1}| &\propto \left| \max(0, \mathbf{t}_i \cdot \mathbf{l}) - \max(0, \mathbf{t}_{i+1} \cdot \mathbf{l}) \right|.
\end{aligned}
\end{equation}
where $I_i$ is the shading intensity of the $i$-th hair segment, $\mathbf{t}_i$ is its tangent direction, and $\mathbf{l}$ is the light direction. 
The shading discontinuity between adjacent Gaussians is represented by $|I_i - I_{i+1}|$, and this value increases with curvature. 
To address this, we introduce a curvature-based blending algorithm that adaptively adjusts the interpolation of Gaussian parameters (color and opacity) based on local strand curvature. For each segment \(i\), we compute the tangent vector \(\mathbf{t}_i\), curvature \(\kappa_i\), and normalized curvature \(\tilde{\kappa}_i\), which is used as the blending weight \(w_i\):
\begin{equation}\small
\begin{aligned}
\mathbf{t}_i &= \frac{\mathbf{p}_{i+1} - \mathbf{p}_i}{\|\mathbf{p}_{i+1} - \mathbf{p}_i\|},\quad \kappa_i = \|\mathbf{t}_i - \mathbf{t}_{i-1}\|, \\
\tilde{\kappa}_i &= \frac{\kappa_i}{\kappa_{\max} + \epsilon}, \quad w_i = \tilde{\kappa}_i,
\end{aligned}
\end{equation}
where \(\mathbf{p}_i\) denotes the 3D position of the \(i\)-th point along the strand, \(\kappa_{\max}\) is the maximum curvature across the strand, and \(\epsilon\) is a small constant for numerical stability.
We apply \(w_i\) to blend spherical harmonics (SH) coefficients and opacity \(\alpha\) between adjacent Gaussians:
\begin{equation}\small
\begin{aligned}
\text{SH}_{\text{blended},i} &= \text{SH}_i \cdot (1 - w_i) + \text{SH}_{i-1} \cdot w_i, \\
\alpha_{\text{blended},i} &= \alpha_i \cdot (1 - w_i) + \alpha_{i-1} \cdot w_i.
\end{aligned}
\end{equation}

As shown in Fig.~\ref{fig:ablation1}, our curvature-based blending significantly improves the visual continuity of hair rendering, particularly in curved regions. This blending is jointly optimized with other Gaussian parameters (e.g., position, scale, SH, opacity) during training, enabling more realistic and temporally coherent appearance in dynamic hair sequences.

\subsection{Inference and Implementation Details}
We train our model on a single A100 GPU using the Adam optimizer and a learning rate of \(1 \times 10^{-4}\) for both stages. In Stage I, each iteration samples 200k points from the hair point cloud. Here we provide the formal definitions for Eq.~\ref{eq:loss_fine}, including the point loss $\mathcal{L}_{\text{point}} = \frac{1}{N}\sum_{i=1}^{N} \lVert \hat{\mathbf{p}}_{i} - \mathbf{p}_{i}^{\mathrm{GT}} \rVert_{2}^{2}$ and the SDF penalty loss $\mathcal{L}_{\text{SDF}} = \frac{1}{N}\sum_{i=1}^{N} \max(0,\,-\,\mathrm{SDF}(\hat{\mathbf{p}}_{i}))$, where $\hat{\mathbf{p}}_{i}$ is the predicted 3D hair point, $\mathbf{p}_{i}^{\mathrm{GT}}$ is the corresponding ground-truth hair point, and $\mathrm{SDF}(\hat{\mathbf{p}}_{i})$ denotes the signed distance to the body surface. We penalize points inside the mesh via the ReLU (i.e., $\max(0,\cdot)$) operation. For Eq.~\ref{eq:loss_fine}, we set \(\lambda_{\text{p}}{=}1.0\) and \(\lambda_{\text{SDF}}{=}0.01\); for Eq.~\ref{eq:loss_render}, we set \(\lambda_{\text{rgb}}{=}1.0\), \(\lambda_{\text{ssim}}{=}0.1\), and \(\lambda_{\text{lpips}}{=}0.1\). 

During inference, we are given a head motion sequence and a canonical hair groom. For \( t = 0 \), we apply only the coarse stage to initialize the dynamic hair. At \( t = 1 \), the fine stage is used with self-attention over the result from \( t = 0 \), and the input flow \( \mathcal{F}_\text{flow}^0 \) is set to 0. For \( t > 1 \), the pipeline operates recurrently as described.
Once the dynamic hair sequence is obtained, we infer hair appearance based on the tracked hair positions to achieve dynamic hair rendering. For further implementation details (runtime and memory analysis), please refer to the Appendices.

%% file: sec/4_experiment.tex
\section{Experiments}
\label{sec:experiment}

\begin{figure*}[htb]
\includegraphics[width=\textwidth]{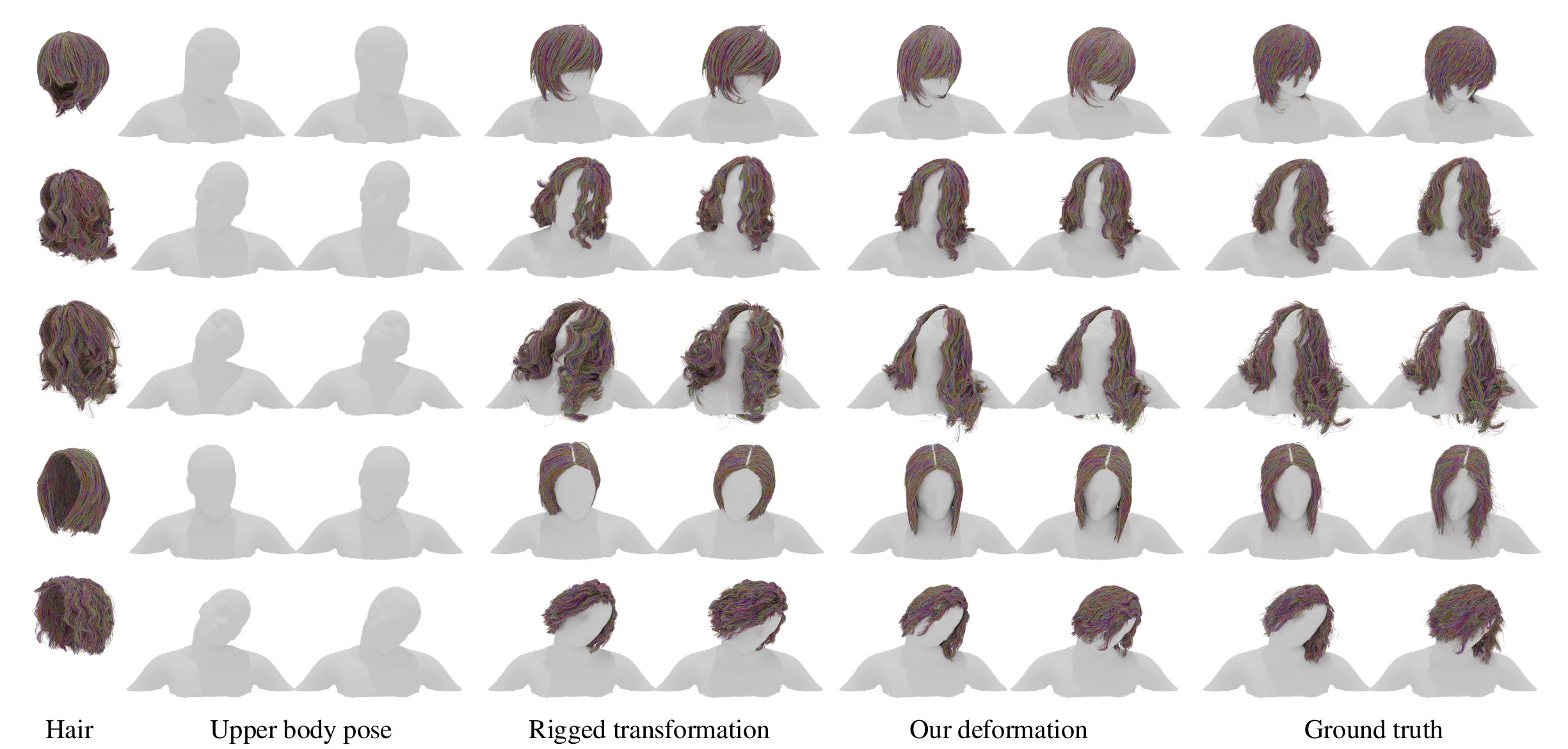}
\caption{\textbf{Hair Deformation Comparison.} From left to right column, given a canonical groom and upper body pose, rigid transformations result in unrealistic hair deformation with upper-body penetration, while our method achieves natural deformation across different grooms with correct collisions.}
\label{fig:deformation}   
\end{figure*}

\begin{figure*}[tb]
\includegraphics[width=\linewidth]{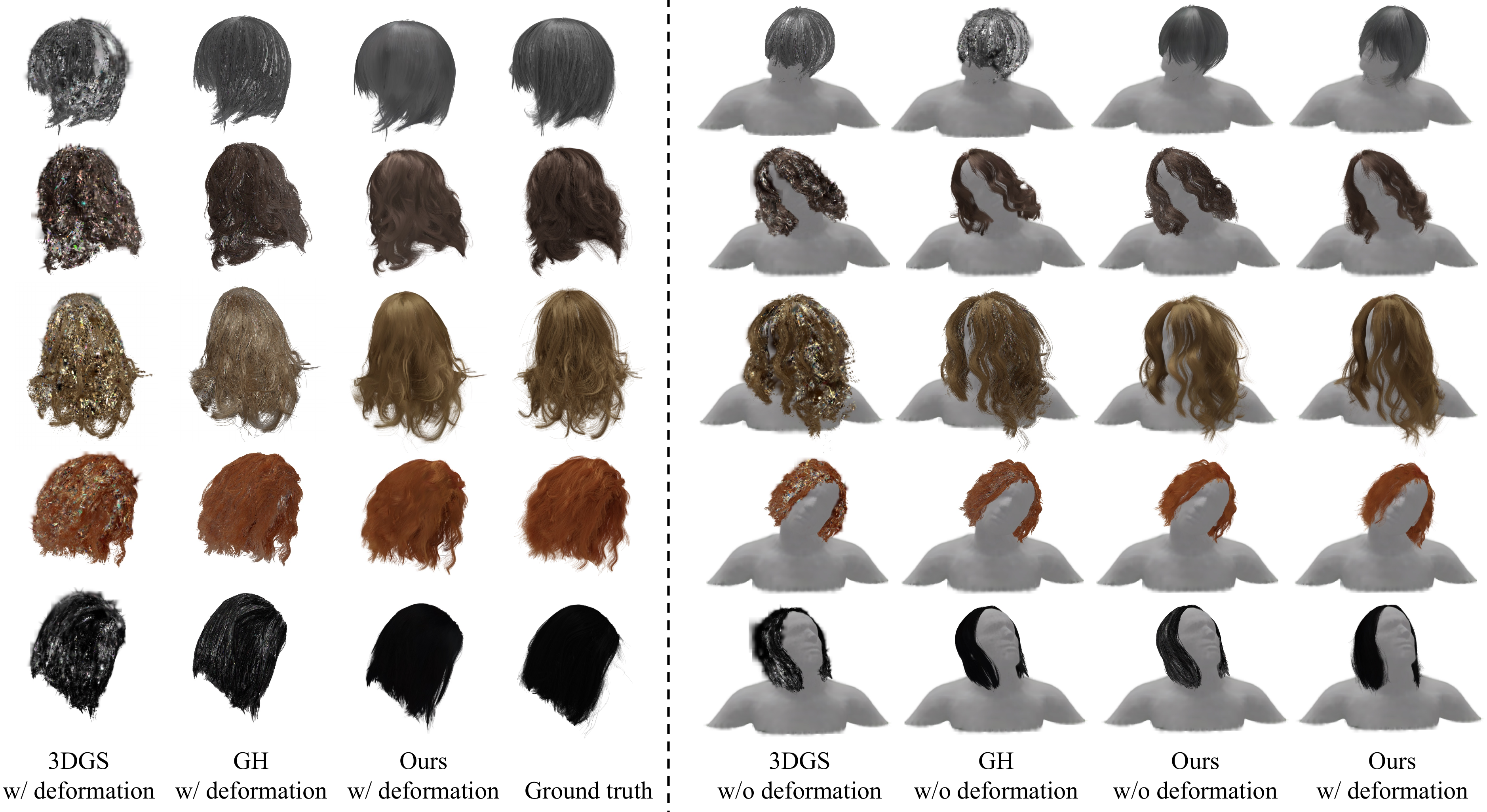}
\vspace{-5mm}
\caption{\textbf{Dynamic hair appearance evaluation.} The left section compares dynamic hair rendering using our hair deformation/tracking model against baseline methods: 3DGS~\cite{kerbl20233d} and Gaussian Haircut (GH)~\cite{zakharov2024human}. The right section shows rendering results merging hair and body Gaussian primitives. Without our hair tracking model, the hair appears rigid and unrealistic, while our appearance model enhances hair rendering quality.}
\vspace{-6mm}
\label{fig:compare_app}
\end{figure*}

\begin{table*}[t]
\centering
\footnotesize
\scalebox{1}{
\begin{tabular}{c||p{1.3cm}<{\centering\arraybackslash}p{0.825cm}<{\centering\arraybackslash}p{0.85cm}<{\centering\arraybackslash}|p{0.85cm}<{\centering\arraybackslash}p{0.85cm}<{\centering\arraybackslash}p{0.85cm}<{\centering\arraybackslash}|p{0.85cm}<{\centering\arraybackslash}p{0.85cm}<{\centering\arraybackslash}p{0.85cm}<{\centering\arraybackslash}} 
  \hline
  \multicolumn{1}{c||}{} & \multicolumn{3}{c|}{\textbf{Ours}} & \multicolumn{3}{c|}{\textbf{Gaussian Haircut~\cite{zakharov2024human}}} & \multicolumn{3}{c}{\textbf{3D GS~\cite{kerbl20233d}}} \\
  \hline
  \textbf{Subject} & \textbf{PSNR}$\uparrow$ & \textbf{SSIM}$\uparrow$ & \textbf{LPIPS}$\downarrow$ 
  & \textbf{PSNR}$\uparrow$ & \textbf{SSIM}$\uparrow$ & \textbf{LPIPS}$\downarrow$ 
  & \textbf{PSNR}$\uparrow$ & \textbf{SSIM}$\uparrow$ & \textbf{LPIPS}$\downarrow$ \\
  \hline
  Subject 1 & \cellcolor{YellowGreen}28.026 & \cellcolor{YellowGreen}0.906 & \cellcolor{YellowGreen}0.101 & 23.248 & 0.873 & \cellcolor{YellowGreen}0.101 & 20.747 & 0.852 & 0.132 \\
  Subject 2 & \cellcolor{YellowGreen}24.817 & \cellcolor{YellowGreen}0.820 & \cellcolor{YellowGreen}0.169 & 20.921 & 0.791 & 0.180 & 19.972 & 0.772 & 0.215 \\
  Subject 3 & \cellcolor{YellowGreen}24.987 & \cellcolor{YellowGreen}0.744 & \cellcolor{YellowGreen}0.227 & 20.960 & 0.742 &  0.236 & 19.972 & 0.687 & 0.287 \\
  Subject 4 & \cellcolor{YellowGreen}27.534 & \cellcolor{YellowGreen}0.955 & \cellcolor{YellowGreen}0.071 & 24.181 & 0.906 & 0.078 & 20.246 & 0.894 & 0.101 \\
  Subject 5 & \cellcolor{YellowGreen}29.681 & \cellcolor{YellowGreen}0.933 & 0.069 & 26.053 & 0.924 & \cellcolor{YellowGreen}0.058 & 23.605 & 0.906 & 0.083 \\
  \hline
  Average & \cellcolor{YellowGreen}27.009 & \cellcolor{YellowGreen}0.871 & \cellcolor{YellowGreen}0.127 & 23.073 & 0.847 & 0.131 & 20.908 & 0.822 & 0.164 \\
  \hline
\end{tabular}
}
\caption{\textbf{Comparison with others baselines on hair appearance.} We report PSNR$\uparrow$, SSIM$\uparrow$, and LPIPS$\downarrow$ for each groomed subject to compare our method with other baselines in rendering quality.}
\label{table:compare_app}
\vspace{-3mm}
\end{table*}

\textbf{Dataset.}
Due to the lack of real hair tracking data, we generate a synthetic dataset using XPBD-based physics simulation~\cite{macklin2016xpbd}. It includes dynamic sequences across diverse hairstyles, driven by motion-captured head movements.
Each simulated groom consists of $1500k$ hair strand with $24$ vertices per strand. For each hairstyle, we simulate $100$ motion sequences of $100$ frames, totaling $10k$ frames. Each frame includes deformed hair positions, head motion parameters, and the upper-body mesh.
To generate our dynamic hair appearance dataset, we simulate $500$ frame sequences per groom and render multi-view videos from $24$ camera angles in Blender. See the Appendices for dataset details.

\begin{wraptable}{r}{0.6\linewidth}
  \centering
  \footnotesize
  \begin{tabular}{c||>{\centering\arraybackslash}p{1.25cm}>{\centering\arraybackslash}p{1.2cm}|>{\centering\arraybackslash}p{1.2cm}>{\centering\arraybackslash}p{1.2cm}} 
    \hline
    \multicolumn{1}{c||}{} & \multicolumn{2}{c|}{\textbf{Ours}} & \multicolumn{2}{c}{\textbf{Rigged hair}} \\
    \hline
    \textbf{Subject} & \textbf{Error}$\downarrow$ & \textbf{Chamfer}$\downarrow$ & \textbf{Error}$\downarrow$ & \textbf{Chamfer}$\downarrow$ \\
    \hline
    Subject 1 & 0.0738 & 0.0233 & 0.1411 & 0.0382 \\
    Subject 2 & 0.0998 & 0.0294 & 0.1911 & 0.0474 \\
    Subject 3 & 0.1187 & 0.0342 & 0.2891 & 0.0637 \\
    Subject 4 & 0.0679 & 0.0260 & 0.1054 & 0.0334 \\
    Subject 5 & 0.0562 & 0.0201 & 0.0928 & 0.0296 \\
    \hline
    Average & \cellcolor{YellowGreen} 0.0832 & \cellcolor{YellowGreen} 0.0266 & 0.1639 & 0.0424 \\
    \hline
  \end{tabular}
  \caption{\textbf{Hair deformation comparison.} $L_2$ and Chamfer distances across normalized subjects.}
  \label{table:compare_geo}
\end{wraptable}

\noindent\textbf{Evaluation.}
We evaluate our method on deformation and appearance, comparing it with baselines across $5$ hair subjects using our synthetic hair dataset. Each groom's training dataset consists of $90$ motion sequences, while testing is performed on the remaining $10$ sequences. For dynamic appearance, we assess unseen hair motions ($100$ frames per subject) and novel views, capturing $100$ views via horizontal camera rotation. See the Appendices for more dataset and evaluation details.

\noindent\textbf{Metrics.} We evaluate the quality of rendered images via reconstruction fidelity (PSNR~\cite{hore2010image}), local structural similarity (SSIM~\cite{hore2010image}), and perceptual similarity (LPIPS~\cite{zhang2018perceptual}) between the synthesized images and the ground truth. We evaluate per-frame motion via the $L2$ error, Chamfer distance~\cite{fan2017point}, between our dense hair point cloud and the ground truth. We evaluate the temporal consistency of the hair motion via the $L2$ error between the flow vectors in the predicted and ground-truth point cloud sequences, and we show the flow error of different settings in Fig.~\ref{fig:ablation3}

\noindent\textbf{Baselines.} Our framework learns dynamic hair motion and time-varying appearance using Gaussians in a differentiable manner. We compare against 3DGS~\cite{kerbl20233d} and Gaussian Haircut~\cite{zakharov2024human}, retraining both on our synthetic dataset for each static hairstyle. Since dynamic hair modeling is underexplored, we integrate our dynamics model into each baseline, optimize their canonical hair appearance, and re-animate hair using our estimated motion for fair comparison. We further benchmark dynamics against two references: rigid-transformed canonical hair (lower bound) and physics-based XPBD~\cite{macklin2016xpbd} results (upper bound). Full baseline-training and implementation details are provided in the Appendices.
\begin{table}[htb]
\centering
\begin{minipage}[t]{0.48\linewidth}
  \centering
  \footnotesize
  \scalebox{1.0}{
    \begin{tabular}{p{3.4cm}||>{\centering\arraybackslash}p{1.8cm}} 
      \hline
      \textbf{Setting} & \textbf{Error}$\downarrow$ \\
      \hline
      Ours w/o SDF & 0.1269 \\
      Ours w/o motion & 0.0964 \\
      Ours w/o atten. & 0.0909 \\
      Ours full & \cellcolor{YellowGreen} 0.0832 \\
      \hline
    \end{tabular}
  }
\caption{\textbf{Hair dynamics ablation.} We show the $L2$ error across settings on deformed hair test sets.}
  \label{table:ablation_geo}
\end{minipage}
\hfill
\begin{minipage}[htb]{0.5\linewidth}
  \centering
  \footnotesize
  \scalebox{0.95}{
    \begin{tabular}{p{2.75cm}||p{0.78cm}<{\centering\arraybackslash}|p{0.78cm}<{\centering\arraybackslash}|p{0.78cm}<{\centering\arraybackslash}} 
      \hline
      \textbf{Setting} & \textbf{PSNR}$\uparrow$ & \textbf{SSIM}$\uparrow$ & \textbf{LPIPS}$\downarrow$ \\
      \hline
      Ours w/o tan. \& blend & 20.89 & 0.80 & 0.19 \\
      Ours w/o blend & 25.08 & 0.88 & \cellcolor{YellowGreen} 0.18 \\
      Ours full & \cellcolor{YellowGreen} 28.12 & \cellcolor{YellowGreen} 0.90 & 0.19 \\
      \hline
    \end{tabular}
  }
\caption{\textbf{Hair appearance ablation.} We compare different appearance model settings on rendering quality.}
\label{table:ablation_app}
\end{minipage}
\end{table}
\vspace{-5mm}
\begin{figure*}[htb]
\includegraphics[width=\textwidth]{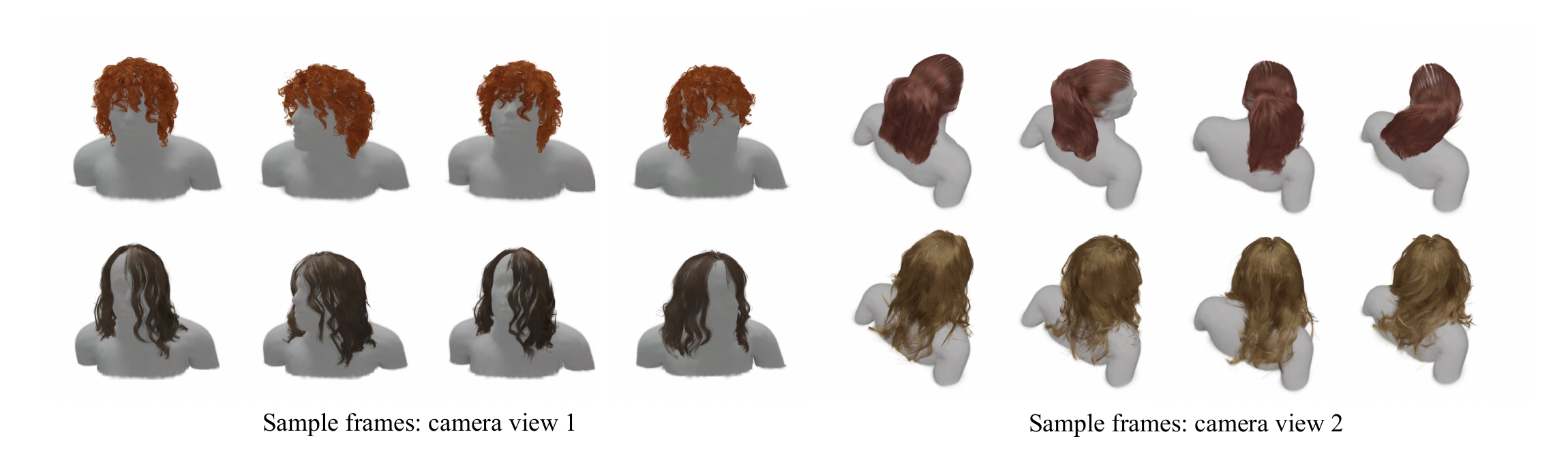}
\caption{\textbf{Dynamic Gaussian Hair qualitative evaluation.} We present qualitative visual results across various hairstyles and camera views, 
where the driven motion is obtained from the Mixamo MoCap data~\cite{mixamo}.}
\label{fig:real_motion} 
\vspace{-5mm}
\end{figure*}
\subsection{Results}
 Tab.~\ref{table:compare_geo} shows that our deformation model significantly enhances realistic hair deformations and reduces errors compared to rigid transformations on static hair. As shown in Fig.~\ref{fig:deformation}, applying rigid transformations to static hair leads to unnatural deformations, such as the absence of gravity effects and hair penetrating the body mesh. Tab.~\ref{table:compare_app} presents a quantitative comparison of our method with GH (Gaussian Haircut)~\cite{zakharov2024human} and 3DGS~\cite{kerbl20233d} on our synthetic dynamic hair dataset. Our approach consistently achieves the highest PSNR and SSIM across all subjects, indicating high-fidelity rendering quality. In Fig.~\ref{fig:compare_app} and~\ref{fig:real_motion} we further demonstrate dynamic hair appearance across different frames and hairstyles. Qualitative comparisons demonstrate that without our appearance optimization, both 3DGS and strand-based Gaussian representations suffer from degraded rendering quality due to hair self-occlusions. In contrast, our motion-dependent appearance model is robust to occlusions, ensuring realistic rendering, while our hair deformation model can further enhance dynamic hair realism. Our dynamic Gaussian hair representation is flexible, allowing integration with body Gaussian primitives for avatar reanimation and high-quality rendering.
Additional visual results are provided in the supplemental video.
\vspace{-2mm}
\subsection{Ablation Study} 
\begin{figure}[t]
\centering
\begin{minipage}[t]{0.48\linewidth}
    \centering
    \includegraphics[width=\linewidth]{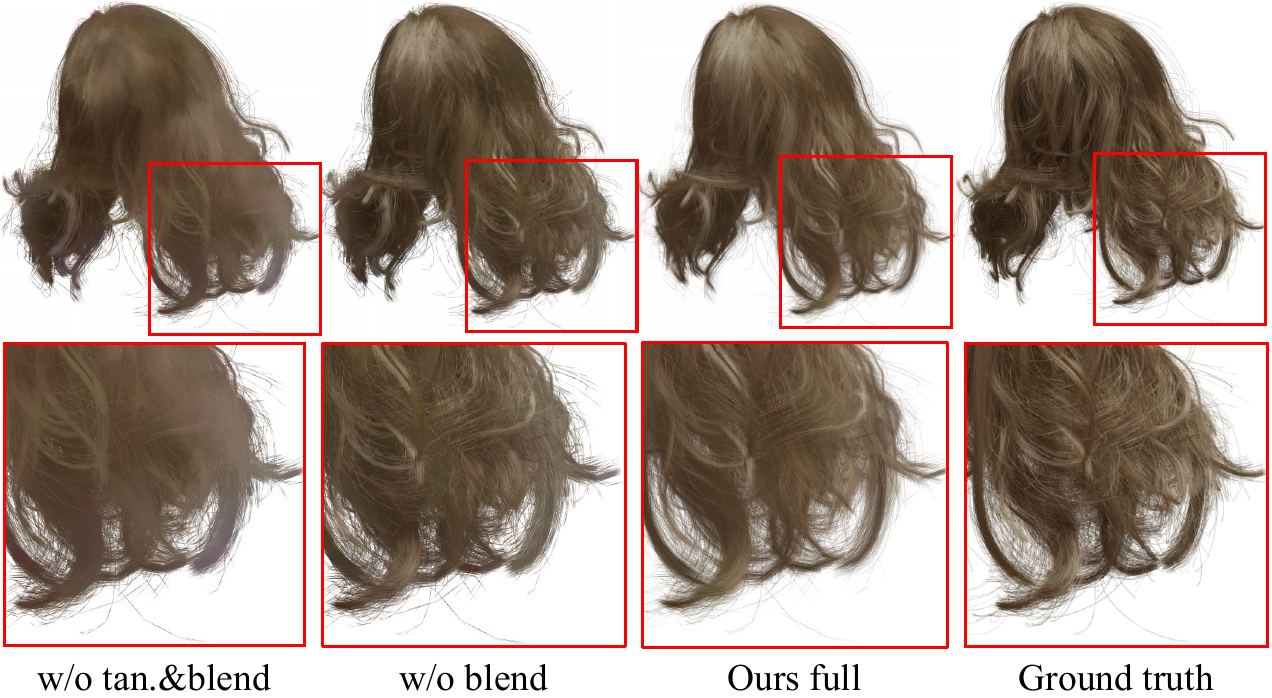}
    \caption{\textbf{Ablation on appearance.} Top: results of appearance model variants. Bottom: zoomed details.}
    \label{fig:ablation1}
\end{minipage}
\hfill
\begin{minipage}[t]{0.48\linewidth}
    \centering
    \includegraphics[width=\linewidth]{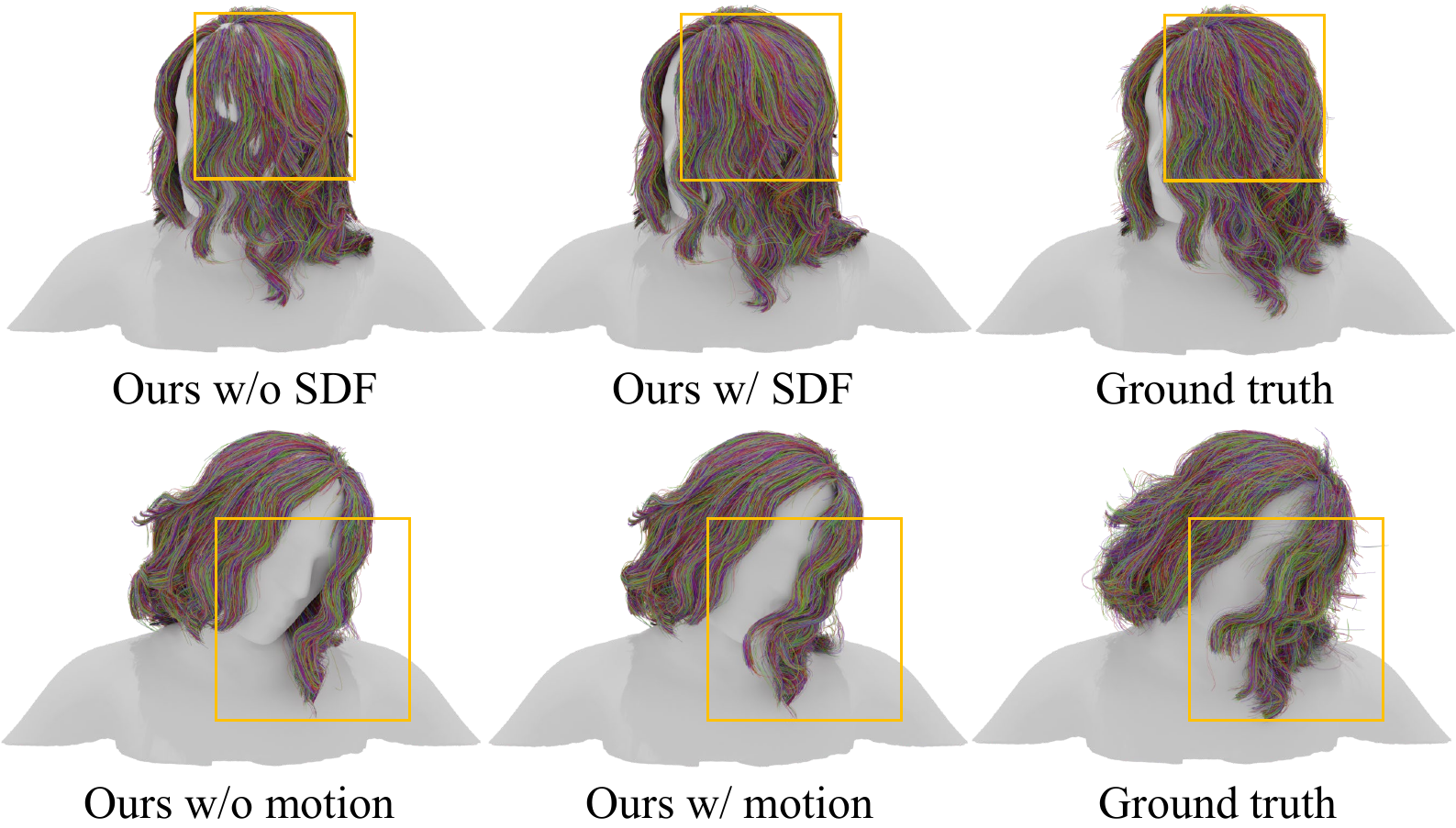}
    \caption{\textbf{Ablation on dynamics.} Top: hair w/wo SDF. Bottom: motion model effect.}
    \label{fig:ablation2}
\end{minipage}
\vspace{-5mm}
\end{figure}

\noindent $\bullet$ \textbf{\textit{Dynamic Hair Modeling.}} Fig.~\ref{fig:ablation2} shows how our proposed approaches impact hair deformation and dynamics. \textbf{Ours w/ SDF} uses our hair-to-body mesh SDF constraint during training, reducing hair penetration into the body mesh. \textbf{Ours w/ motion} shows enhanced hair dynamics effects, such as inertia and gravity, due to the hair motion model. \textbf{Ours w/o SDF} removes the SDF constraint, \textbf{Ours w/o motion} uses only the coarse hair deformation model, and \textbf{Ours w/o atten} removes the cross-attention component in our fine stage. These results indicate that our full coarse-to-fine approach achieves the best quality. Detailed quantitative results in Tab.~\ref{table:ablation_geo} confirm this. 

\begin{wrapfigure}{r}{0.5\linewidth}
  \centering
  \vspace{-3mm}
  \includegraphics[width=\linewidth]{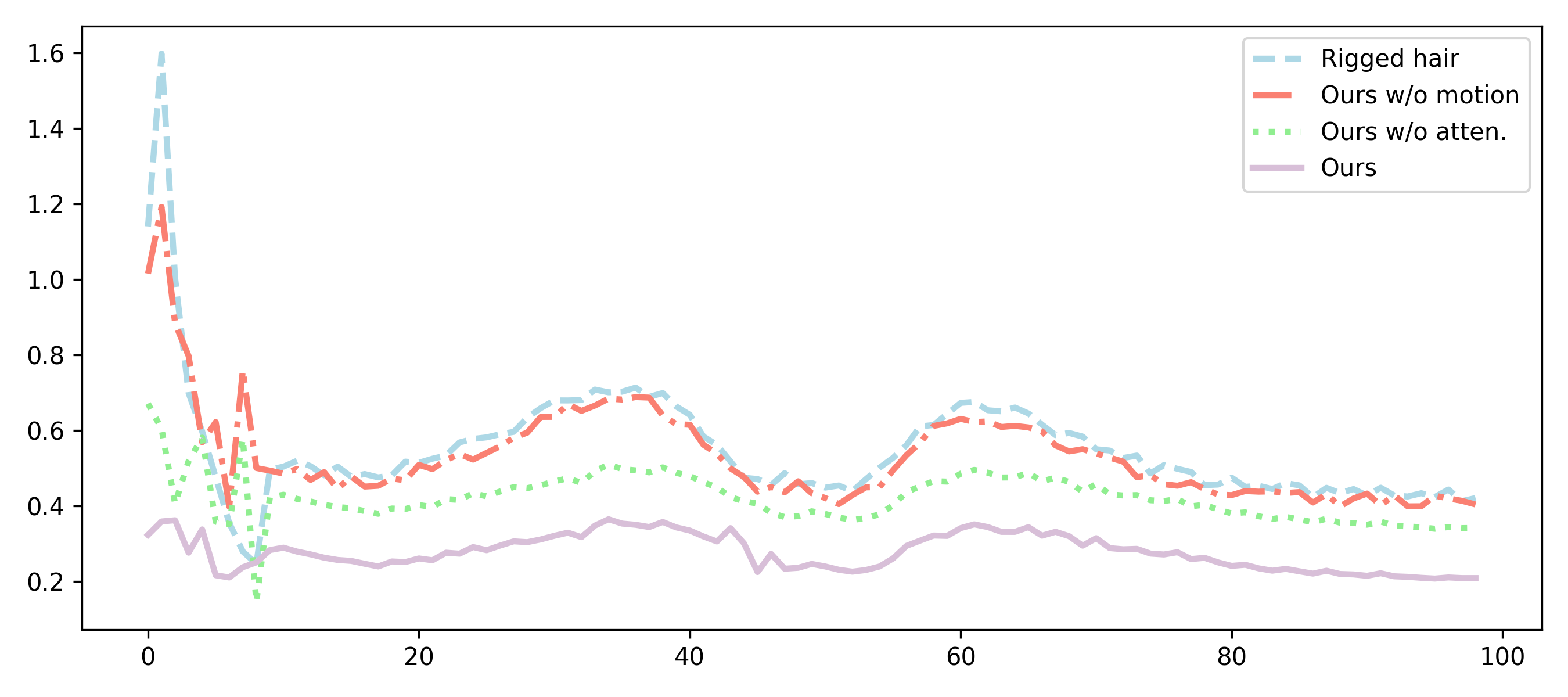} 
  \vspace{-4mm}
  \caption{\textbf{Hair dynamics ablation.} Flow error (\(\times 10^{-3}\)) per timestep for different settings.}
  \label{fig:ablation3}
\end{wrapfigure}
To evaluate temporal consistency, Fig.~\ref{fig:ablation3} plots the flow vector error across a sequence for different settings. While our coarse stage shows similar performance to rigid hair in terms of dynamics, the fine model significantly improves motion effects. Additionally, the attention module further enhances hair dynamics and temporal consistency.
\noindent $\bullet$ \textbf{\textit{Dynamic Hair Appearance Optimization}}. Fig.~\ref{fig:ablation1} compares different configurations of our appearance model: Ours without tangent features or curvature-based blending (\textbf{w/o tan.\&blend}), ours without curvature-based blending (\textbf{w/o blend}), and our full model (\textbf{Ours full}).
It demonstrates that conditioning on structural features like hair tangent vectors enhances local detail and sharpness, while curvature-based blending further improves smoothness and scattering between hair segments, creating more realistic transitions. 
As indicated in Tab.~\ref{table:ablation_app}, incorporating tangent features improves natural hair perception, while curvature blending significantly enhances rendering quality, resulting in more visually appealing hair.
\vspace{-3mm}

%% file: sec/5_conclusion.tex
\vspace{-2mm}
\section{Conclusion}\label{sec:conclusion}
We propose \textbf{Dynamic Gaussian Hair} (\dgh), a novel data-driven framework for dynamic hair generation with animatable 3D Gaussians.
\dgh models a wide range of hairstyle deformation driven by arbitrary head motions (including long hair, curly hair, ponytails, etc.), handling dynamics and hair-body collision, within a coarse-to-fine strategy.
Hair appearance is represented by motion-dependent connected Gaussians to handle variations under intricate motions at render time, enabling high-fidelity dynamic hair modeling and rendering. Future work and limitations are addressed in the Appendices.
\vspace{-2mm}
\section{Acknowledgment}\label{sec:acknowledgment}
We sincerely thank Gene Lin for his help and discussions regarding the synthetic data generation pipeline, and we also appreciate Yu Ding for his assistance with dynamic hair and body merging.

%% file: sec/X_suppl.tex
\appendixtitle
In this document, we provide more details for the dataset, method, experiments, and more qualitative results, as an extension of Sec. 3 and Sec. 4 in the main paper. Please also refer to the video demo for dynamic hair results, comparison, ablation study, and more results. 

\section{Dataset Details}
As mentioned in the Sec. 1, due to the lack of real captures of dynamic hair deformations with accurate strand tracking, we create a new synthetic dataset capturing both hair geometry and appearance. The hair dataset breakdown is shown below:
in Fig.~\ref{fig:hairstyle}, we present a variety of hairstyles in our synthetic hair dataset, including curly, wavy, blowout, ponytail and etc, and we show the distributions of hairstyle types and hair lengths.
\begin{figure*}[htb]
\centering
\includegraphics[width=\textwidth]{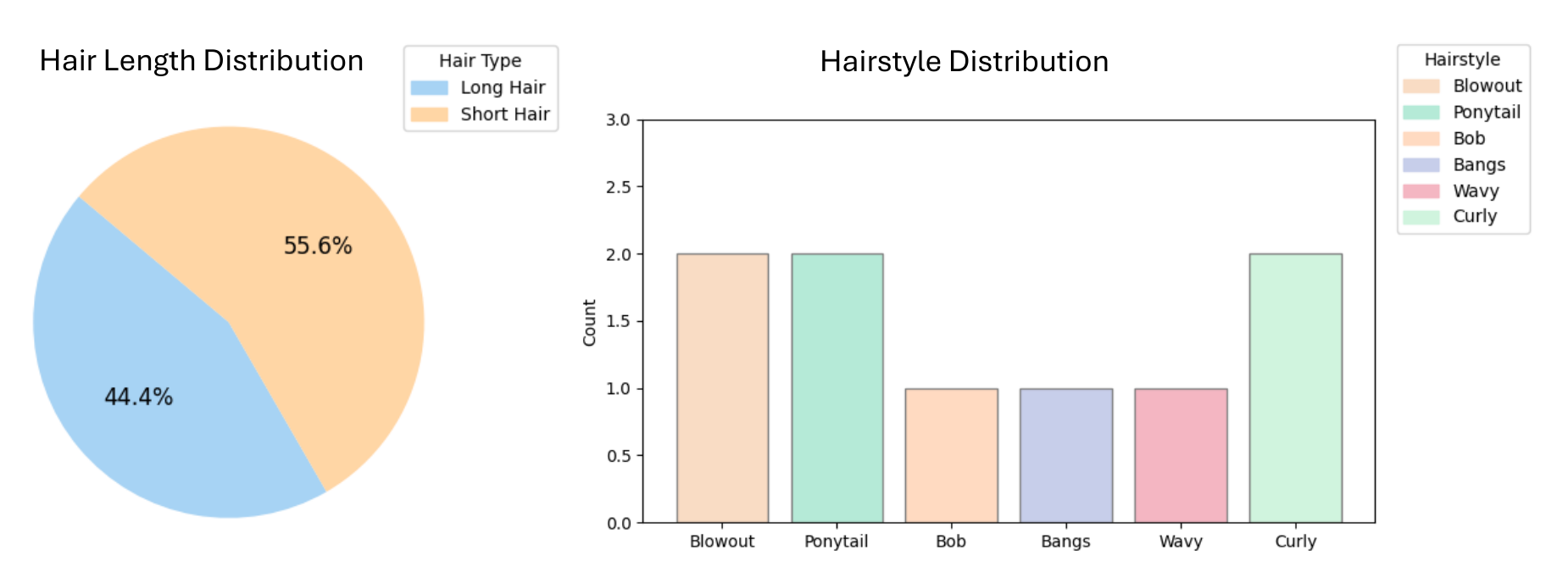}
\caption{\textbf{Hair length and hairstyle distribution}}
\label{fig:hairstyle}
\end{figure*}
\subsection{Geometry Dataset (Sec. 4)}
\begin{figure*}[htb]
\centering
\includegraphics[width=\textwidth]{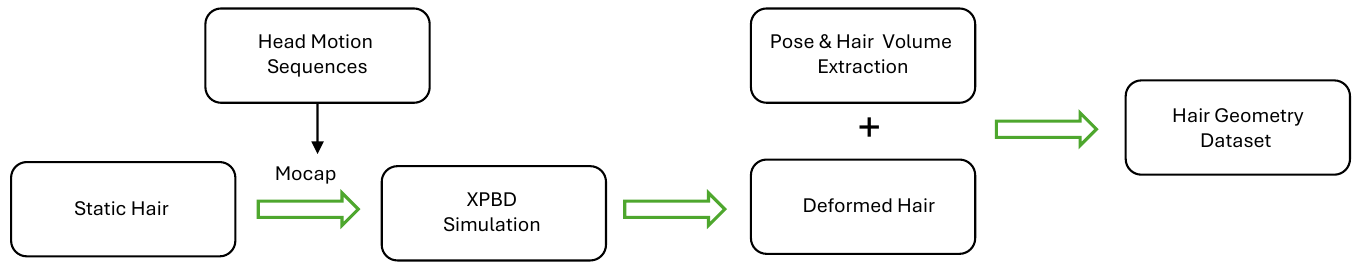}
\caption{\textbf{Hair Geometry Dataset Generation Pipeline}}
\label{fig:geo_data_pipeline}
\end{figure*}

To capture head motions, we rotate the head to record head movements. Instead of linear speed between two head positions, we apply spline-based interpolation to simulate varying motion speeds. This approach allows us to more effectively capture secondary hair motions when simulating the hair with head motions with realistic damping and inertial dynamics. 
For each hairstyle, we animated with XPBD-based physics simulation~\cite{macklin2016xpbd} driven by captured head motions, and for each hairstyle, we generate 100 motion sequences, and each motion sequence contains 100 frames. For each frame, we record the head mesh, deformed hair strands, and then post-process to get hair volume and pose volume. We show the hair geometry dataset generation pipeline in Fig.~\ref{fig:geo_data_pipeline}

\begin{figure*}[htb]
\centering
\includegraphics[width=\textwidth]{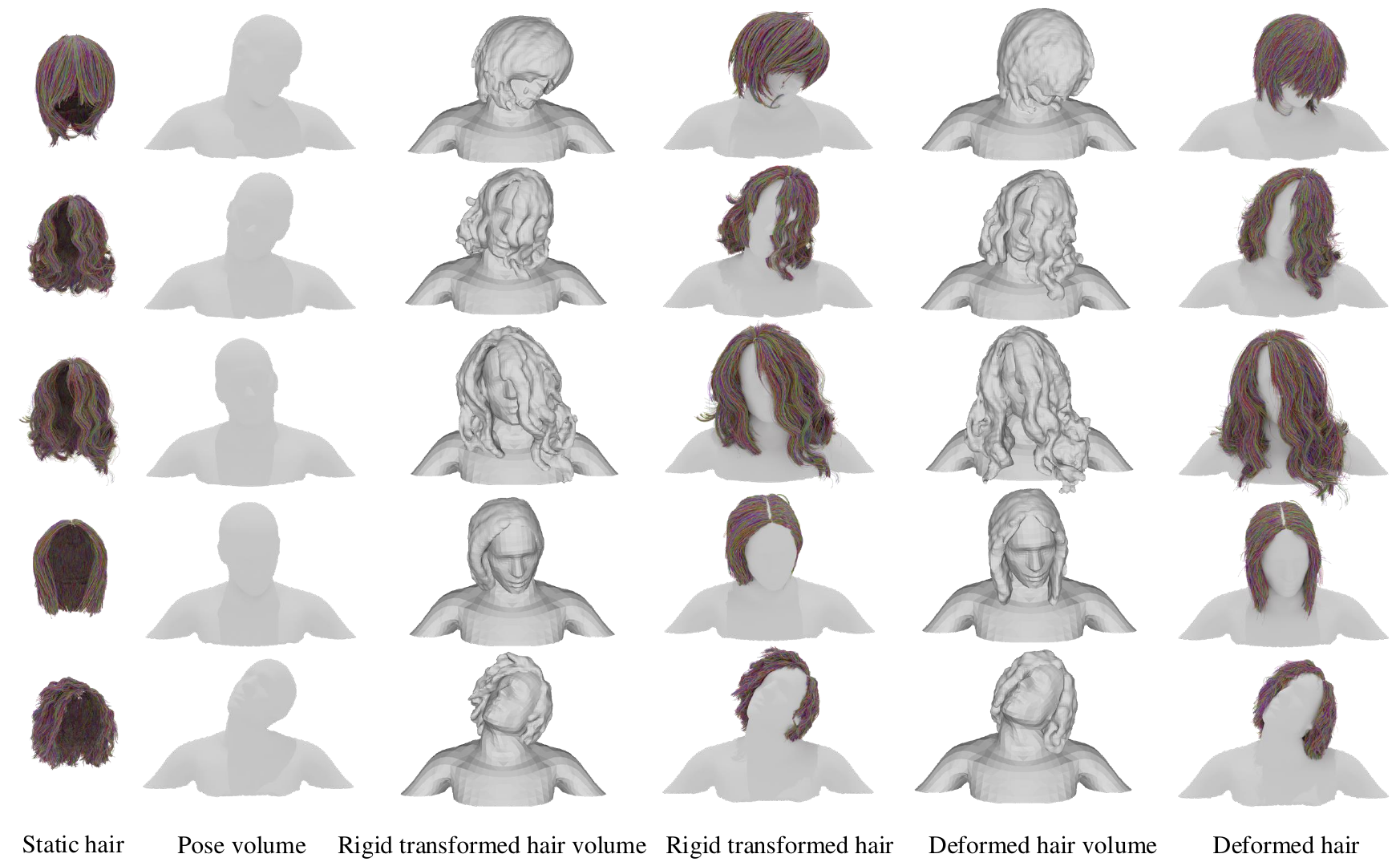}
\caption{Training samples from the hair deformation dataset, including static hair, pose volume, rigidly transformed static hair, recorded hair volume, deformed hair, and corresponding recorded hair volume.}
\label{fig:geo_dataset}
\end{figure*}

Fig.~\ref{fig:geo_dataset} presents samples from our hair deformation dataset, including static hair with strands, upper body SDF volume, static hair SDF volume, deformed hair strands, and deformed hair SDF volumes for each hair groom. 
For hair SDF volumes, hair point cloud (PC) is converted to an SDF grid by computing the distance from each voxel to the nearest point in the PC. 

\textbf{Training dataset.}
We train each hairstyle independently to obtain a hairstyle-specific hair deformation model. For each hairstyle, we use 90 motion subjects, resulting in a geometry training dataset of 9K frames. In the coarse stage, we randomly sample a frame and apply rigid transformations to the static hair using the sampled head pose. The network learns per-point displacements by iteratively sampling different poses. In the fine stage, we learn a flow vector field to deform the hair from frame $t{-}1$ to frame $t$, using randomly sampled points from the deformed hair at time $t{-}1$. Each training batch consists of 200K randomly sampled hair points.

\textbf{Testing dataset.}
We evaluate our dynamic hair model across different hairstyles using 10 motion sequences, resulting in 1K frames. We report the average $L_{2}$ error between the predicted and ground-truth deformed hair, as well as the $L_{2}$ error of the estimated flow vectors (displacements between frames $t{-}1$ and $t$).

\subsection{Appearance Dataset (Sec. 4)}
\begin{figure*}[htb]
\centering
\includegraphics[width=\textwidth]{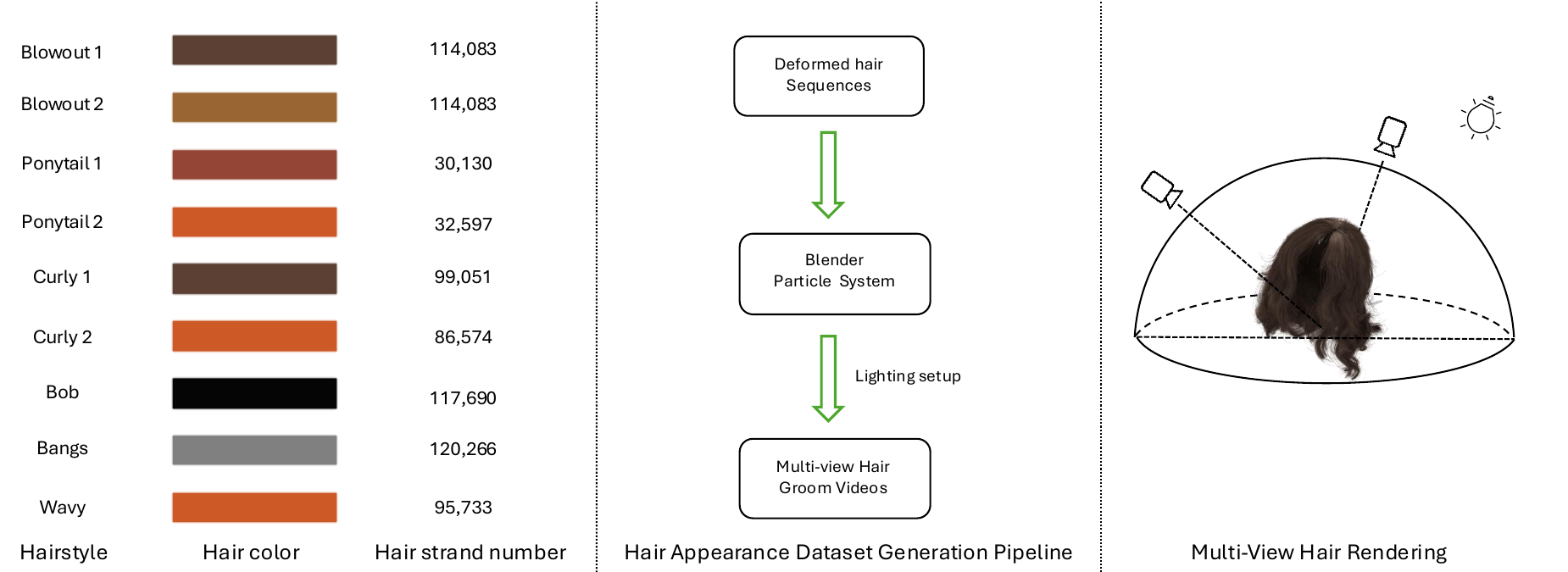}
\caption{\textbf{Hairstyle Color Variations and Hair Appearance Dataset Generation Pipeline}}
\label{fig:color_data_pipeline}
\end{figure*}

\begin{figure*}[htb]
\centering
\includegraphics[width=\textwidth]{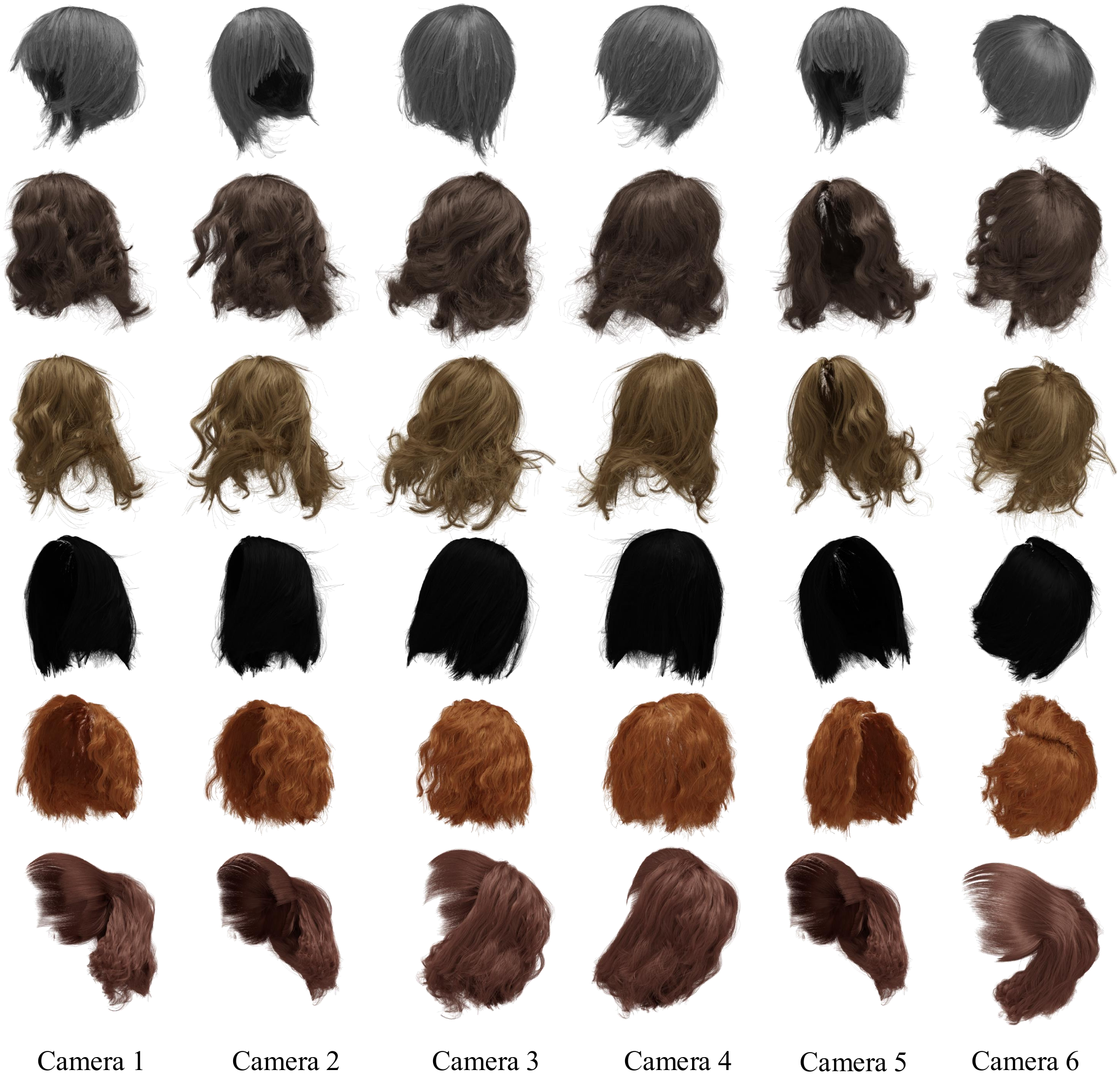}
\caption{Training samples from the appearance dataset, which presents a static frame of various hair grooms and rendering results from six different camera views.} 
\label{fig:render_dataset}
\end{figure*}

Once obtaining the deformed hair sequences, we render multi-view videos of various hair grooms in Blender (particle system) under a consistent lighting setup to train our appearance model. Each frame has a resolution of 1024~$\times$1024. For each hairstyle, we include diverse hair colors. Fig.~\ref{fig:color_data_pipeline} (left) shows the color variations for each hairstyle, (middle) illustrates the hair appearance dataset generation pipeline, and (right) presents a sample setup demonstrating how deformed hair is positioned for multi-view rendering.
Each dynamic frame includes 24 views: 12 cameras evenly distributed horizontally around the object and 12 randomly positioned on the upper semi-sphere to capture diverse angles. Fig.~\ref{fig:render_dataset} illustrates multi-view rendering samples for different hair grooms. 

\textbf{Training dataset}
We train our motion-dependent hair appearance model using multi-view video sequences. Each sequence contains 500 motion frames, and for each frame, we render images from 24 camera viewpoints distributed over a semi-sphere. This setup yields a total of 12K training images per hairstyle.

\textbf{Testing dataset}
To evaluate our appearance model, we generate a testing dataset consisting of 100 frames featuring unseen hair motion. The test sequence is captured using a horizontal camera rotation arc, with each camera looking at the hair center. Viewpoints are sampled across a vertical range of $0^\circ$–$45^\circ$ on the viewing semi-sphere, providing diverse angular coverage and challenging view-dependent appearance variations.

\section{Network and Evaluation Details}
Hair, as a complex and dynamic structure, exhibits intricate spatial patterns and deformations that are challenging to model with lower-dimensional representations, the hair appearance also suffer challenges as the moving pattern caused occlusion. To address this, we propose two stage modeling: which Stage I and II as Hair Dynamics Model and Appearance Model. As summarized in Tab.~\ref{tab:stage}, Stage~I adopts a coarse-to-fine design: the coarse stage learns hair deformation from static hair, while the fine stage refines secondary dynamics 
(e.g., inertia and damping).

\begin{table}[h]
\centering
\vspace{0.5em}
\begin{tabular}{@{}ll@{}}
\toprule
\textbf{Module} & \textbf{Function} \\ \midrule
Stage~I (coarse stage) & Learns hair deformation given head motions \\
Stage~I (fine stage)   & Adds secondary motion refinement \\
Stage~II               & Optimizes dynamic Gaussian hair appearance \\ 
\bottomrule
\end{tabular}
\caption{\textbf{Overview of model stages and their corresponding functions.}}
\label{tab:stage}
\end{table}

For Hair Dynamics Model, we adopt a coarse-to-fine framework, representing static hair as a SDF volume to capture its complete 3D structure and spatial relationships. In this section, we present the architecture of our 3D CNN networks, \( \mathcal{E}_{\text{pose}} \) and \( \mathcal{E}_{\text{hair}} \), the implicit networks \( \mathcal{M} \), \( \mathcal{D} \), and \( \mathcal{D}^* \), along with dataset and evaluation details.

\subsection{Coarse-to-Fine Framework (Sec. 3.1)}
\begin{figure*}[htb]
\centering
\includegraphics[width=\textwidth]{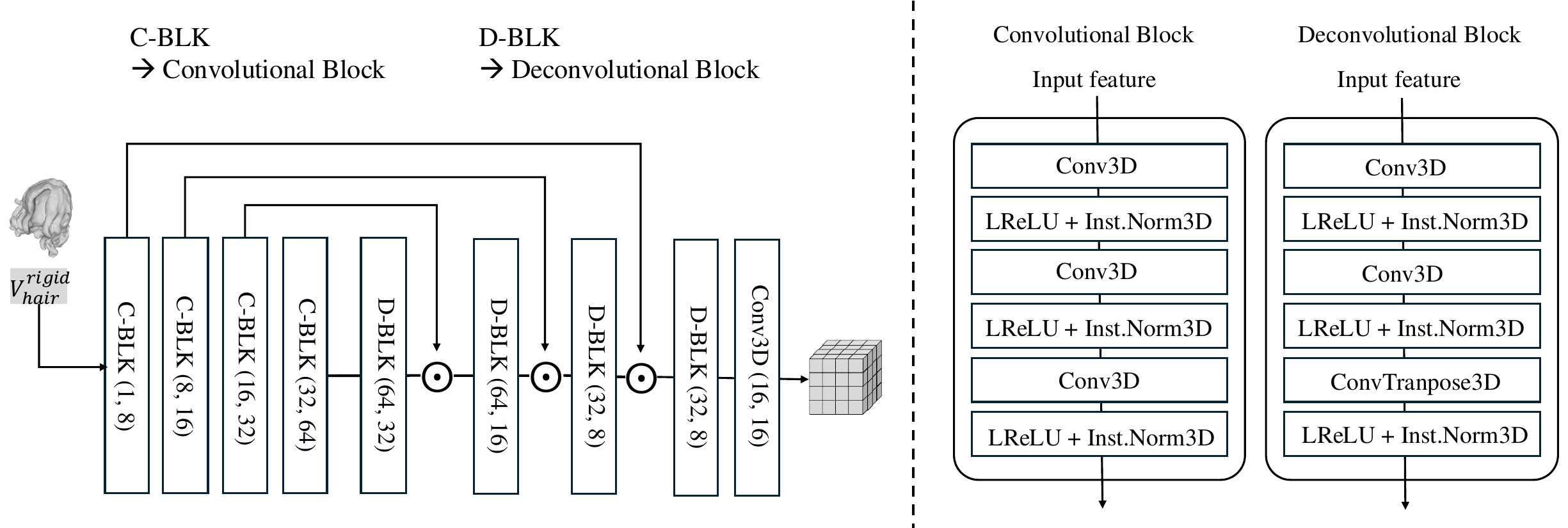}
\caption{Left: Our 3D CNN networks for \( \mathcal{E}_{\text{pose}} \) and \( \mathcal{E}_{\text{hair}} \); Right: Convolutional and deconvolutional blocks.} 
\label{fig:3dcnn}
\end{figure*}

 In the \textbf{Coarse Stage}, we utilize a 3D U-Net architecture for \(\mathcal{E}_{\text{pose}}\) and \(\mathcal{E}_{\text{hair}}\) with input volumes of resolution \(128 \times 128 \times 128\). We present the network structure in Fig.~\ref{fig:3dcnn}. The network employs separate encoders for static hair and pose, each comprising four convolutional blocks with filter sizes \(F = 4\), \(2F\), \(4F\), and \(8F\), utilizing 3D convolutions, LeakyReLU activations, and Instance Normalization to effectively extract hierarchical features.

The decoder mirrors the encoder with four deconvolutional blocks, progressively upsampling and reconstructing volumetric features while reducing the number of filters from \(8F\) to \(F\). Skip connections between corresponding encoder and decoder blocks preserve high-resolution details. With a base filter size of \(F = 4\) and output channels of 16, the 3D U-Net effectively captures global context and fine-grained details of static hair, enabling precise learning of hair deformation by leveraging the static hair structure.

In the \textbf{Fine Stage}, the encoders \(\mathcal{E}_{\text{pose}}\) and \(\mathcal{E}_{\text{hair}}\) retain the same structure, with \(\mathcal{E}_{\text{hair}}\) sharing weights across timesteps during training. The MLPs (\(\mathcal{M}, \mathcal{D}^*\)) consist of six fully connected layers with dimensions \([176, 512, 512, 256, 128, 3]\) and \([239, 512, 512, 256, 128, 3]\), respectively. Residual connections are employed to enhance feature representation by concatenating input features with intermediate outputs. Leaky ReLU is used as the activation function for all layers except the final one, ensuring non-linearity and stability. The final layer outputs a 3-dimensional vector, directly supervised by the ground truth displacement and 3D flow vector.

The fine stage estimates the flow vector from $V^{t-1}_{\text{hair}}$ to $V^{t}_{\text{hair}}$. Since $V^{t}_{\text{hair}}$ is not available, cross-attention~\cite{vaswani2017attention} between $V^{t-1}_{\text{hair}}$ and $V^{t-2}_{\text{hair}}$ infers dynamics. ``Ours w/o atten'' uses only $V^{t-1}_{\text{hair}}$ to estimate the flow vector. In the cross-attention, Q, K, V is $V^{t-2}_{\text{hair}}$, $V^{t-1}_{\text{hair}}$, $V^{t-1}_{\text{hair}}$. This approach leverages the immediate context from $V^{t-1}_{\text{hair}}$ and longer-term dynamics from $V^{t-2}_{\text{hair}}$, enabling the model to capture motion patterns and ensure temporal consistency for a smooth and reliable estimation of $V^{t}_{\text{hair}}$.


\subsection{Dynamic Hair Appearance Network and Implementation(Sec. 3.2)}
For dynamic hair appearance optimization, the encoder \(\mathcal{E}_{\text{hair}}\) encodes the deformed hair to capture global motion features and 3D spatial information. A lightweight MLP \(\mathcal{D}\) with dimensions \([169, 256, 256, 256, 128, 50]\) decodes the Spherical Harmonics coefficients for color, the scale factor along the hair length direction, and opacity. Positional encoding \(E\) for the position \(\mathbf{p}\), tangent vector \(\mathbf{t}\), and view direction \(\mathbf{d}\) follows the encoding function proposed in NeRF~\cite{mildenhall2021nerf}. 
For Gaussian primitive initialization, we make a cylindrical Gaussian primitive attached to each hair segment with a length much greater than its radius. Similar to Gaussian Haircut~\cite{zakharov2024human}, the scale of Gaussian primitives has only one degree of freedom, which is proportional to the length of the line segment, while the other two are fixed to a small predefined value. 




\begin{figure}[b]
\centering
\vspace{-2mm}
\includegraphics[width=\textwidth]{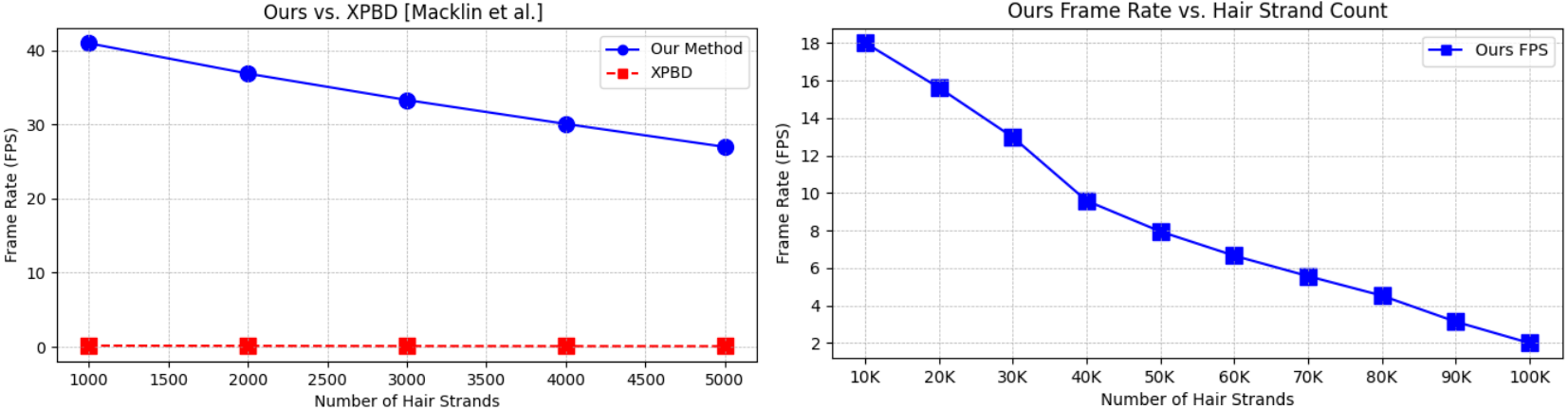}
\vspace{-6mm}
\caption{DGH achieves real-time performance with grooms of 10K strands (unoptimized upper bound). Physics-based simulation (XPBD) runs below 1 FPS with hair interpolation and equal high-quality photoreal rendering in Blender 3D engine.}
\vspace{-3mm}
\label{fig:runtime} 
\end{figure}

\subsection{Evaluation Details (Sec. 4)}
\noindent\textbf{Metrics.}
We evaluate hair motion over time by comparing the flow vectors between consecutive frames in the predicted and ground truth point cloud sequences. The flow vector for each point \( i \) at frame \( t \) in the predicted point cloud is defined as \( \mathcal{F}_{i}^{t} = \mathbf{P}_{i}^{t+1} - \mathbf{P}_{i}^{t} \), where \( \mathbf{P}_{i} \) represents the predicted point cloud positions. Similarly, the flow vector in the ground truth point cloud is defined as \( \hat{\mathcal{F}}_{i}^{t} = \hat{\mathbf{P}}_{i}^{t+1} - \hat{\mathbf{P}}_{i}^{t} \), where \( \hat{\mathbf{P}}_{i} \) represents the GT point cloud positions. The total flow error at each timestep \( t \) over all points in the point cloud is given by \( f^{t}_{\text{error}} = \frac{1}{N} \sum_{i=1}^{N} \left\| \mathcal{F}_{i}^{t} - \hat{\mathcal{F}}_{i}^{t} \right\|_2 \), where \( N \) is the total number of points in the point cloud.

{\renewcommand{\tabcolsep}{3pt}
\begin{table}[h]
    \centering
    \scalebox{1.}{
    \begin{tabular}{|l||c|c|c|c|}
    \hline
    \textbf{Method} & \textbf{PSNR} $\uparrow$ & \textbf{SSIM} $\uparrow$ & \textbf{LPIPS} $\downarrow$\\
    \hline
    Ours w/o LPIPS \& SSIM & 27.82 & 0.8874 & 0.2100 \\
    Ours w/o LPIPS & 28.01 & 0.8923 & 0.1987 \\
    Ours & \textbf{28.12} & \textbf{0.9004} & \textbf{0.1881} \\
    \hline
    \end{tabular}}
    \caption{Ablation of different losses for appearance optimization.}
    \label{tab:app_loss}
\end{table}
}

\noindent\textbf{Inference.}
We train and test our model on a single A100 GPU, with training times of approximately 20 hours for the dynamic hair coarse stage, 20 hours for the fine stage, and 26 hours for the appearance model. During inference, the dynamic hair model achieves 2.0 FPS for approximately 150K hair strands, significantly outperforming the XPBD~\cite{macklin2016xpbd} physics-based simulation engine that is used to generate the synthetic hair dataset, which runs at approximately 0.33 FPS. Meanwhile, dynamic hair novel view synthesis reaches 2.22 FPS, with faster speeds for fewer strands.

\noindent\textbf{Runtime and memory performance analysis.}
DGH design is suitable for low-compute devices. Our implementation is an upper bound. Neural net inference can be accelerated with TensorRT~\cite{nvidia_tensorrt} and additional strands linearly interpolated. For hair deformation runtime analysis, we report runtime with various strand numbers on RTX4090 GPU, compared with XPBD, as shown in Fig~\ref{fig:runtime}. Hair inference required 5GB, while high-quality grooms (150K+ strands) required 20GB. For dynamic hair rendering runtime analysis, our appearance model renders 1K images in 0.45 seconds (2.22 FPS), whereas Blender's hair shader rendering takes 6 seconds per image (0.17 FPS).

\noindent\textbf{Appearance loss ablations.}
We present the appearance loss ablation study in Tab.~\ref{tab:app_loss}. For the final loss constraint, we incorporate both perceptual loss (LPIPS) and structural loss (SSIM), demonstrating that our full method achieves the best performance.


\begin{figure*}[htb]
\centering
\includegraphics[width=\textwidth]{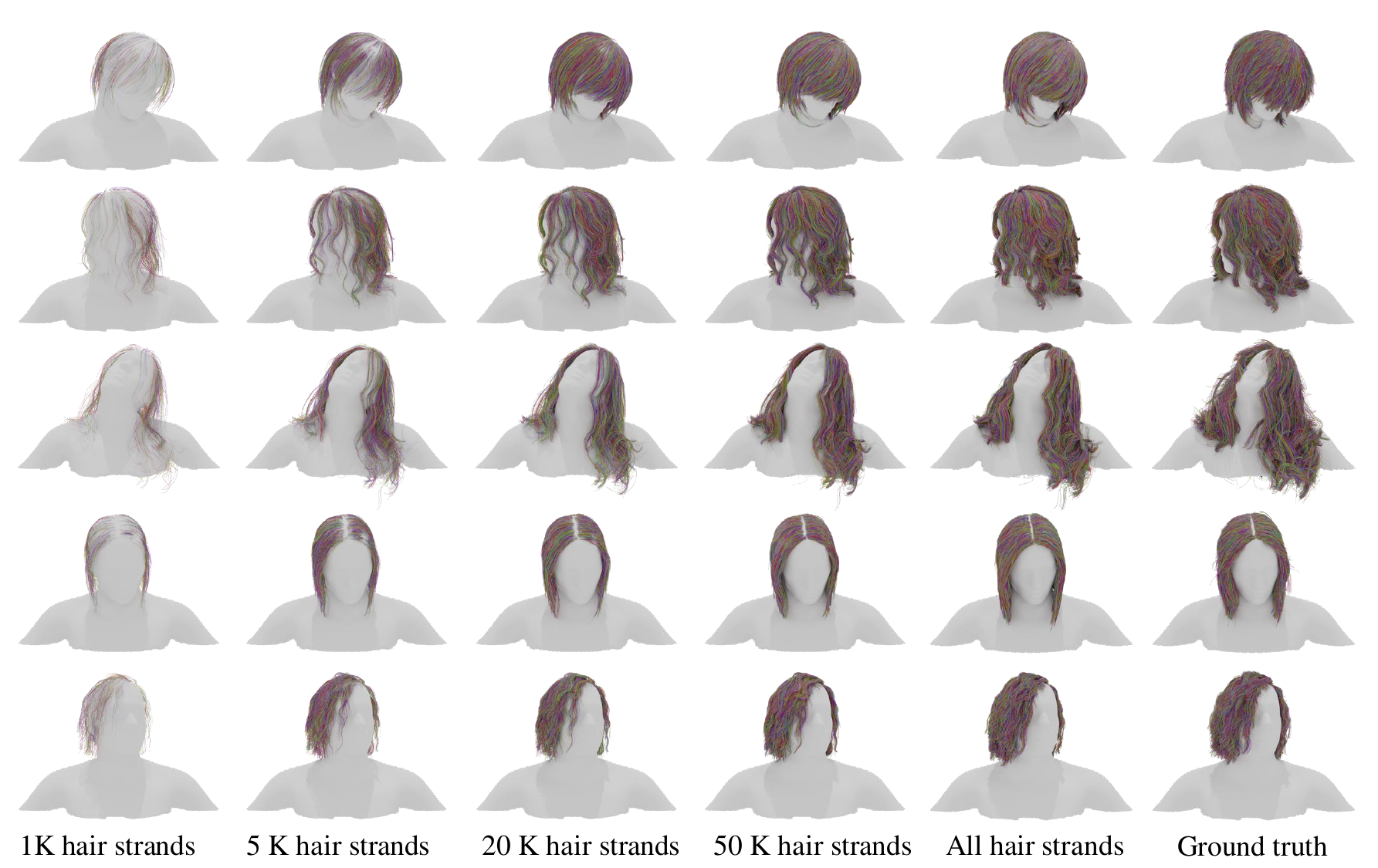}
\caption{Hair deformation with varying strand counts. From left to right: increasing the inferred number of hair strands, totaling approximately 100K to 150K strands.}
\label{fig:hair_strand}
\end{figure*}

\begin{figure*}[htb]
\centering
\includegraphics[width=\textwidth]{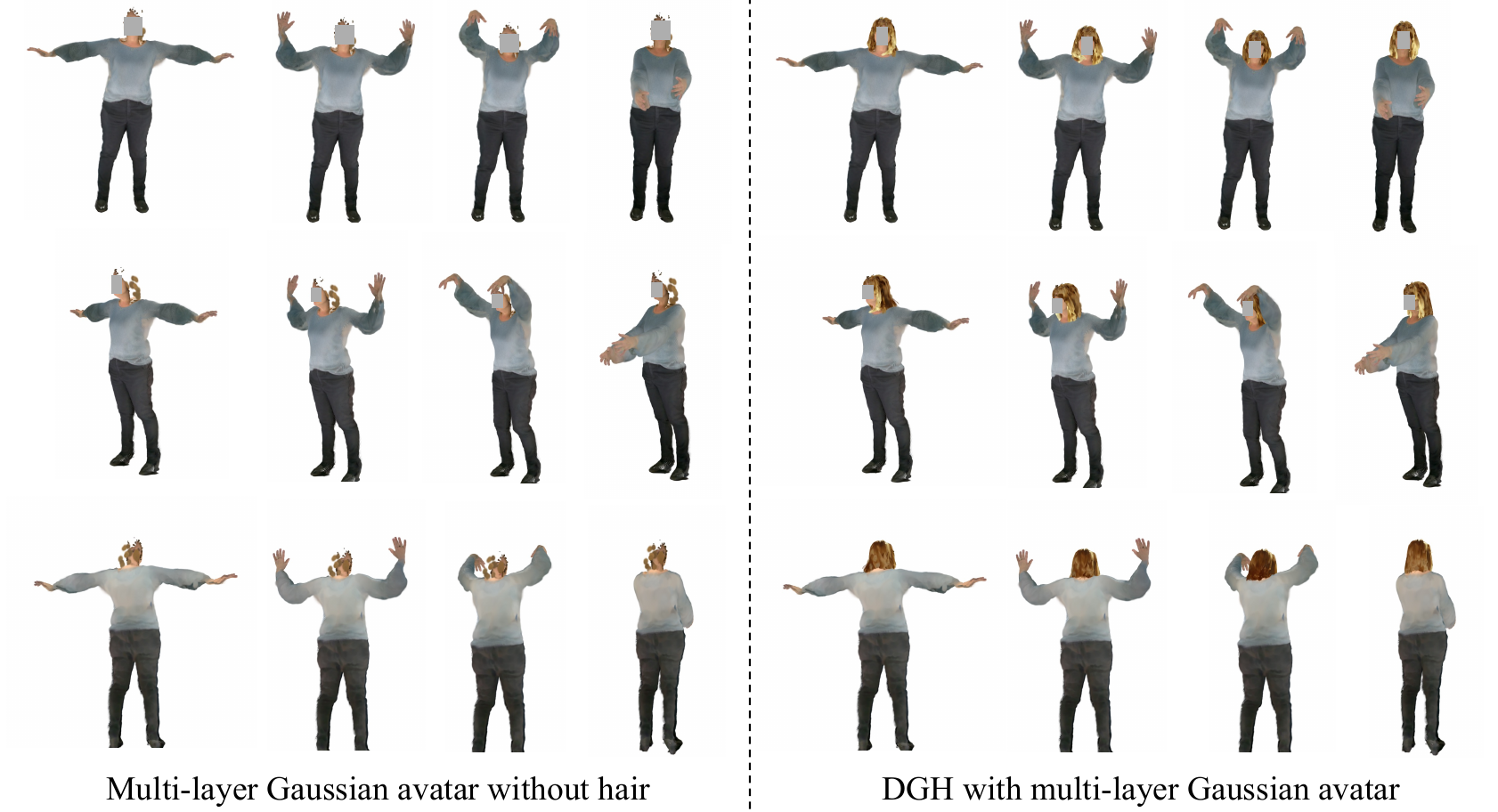}
\caption{Our method merges the dynamic hair layer with a per-frame animatable Gaussian avatar. By employing a multi-layer Gaussian representation (left), we obtain an avatar without the hair layer. Applying DGH within this multi-layer framework enables realistic avatar re-animation (right). \textit{Note: The human face is masked to comply with privacy regulations.}}
\label{fig:real_human}
\end{figure*}

\section{Qualitative Results}
\subsection{Hair Deformation.}
In Fig.~\ref{fig:hair_strand}, we present the hair deformation test on arbitrary head poses and various hairstyles. We also present hair deformation results tested with varying numbers of hair strands, ranging from 1K to the full set of approximately 150K strands. Our volumetric implicit function robustly infers hair deformation for arbitrary strand numbers without requiring post-processing or hair interpolation. 

\begin{figure*}[htb]
\centering
\includegraphics[width=0.9\textwidth]{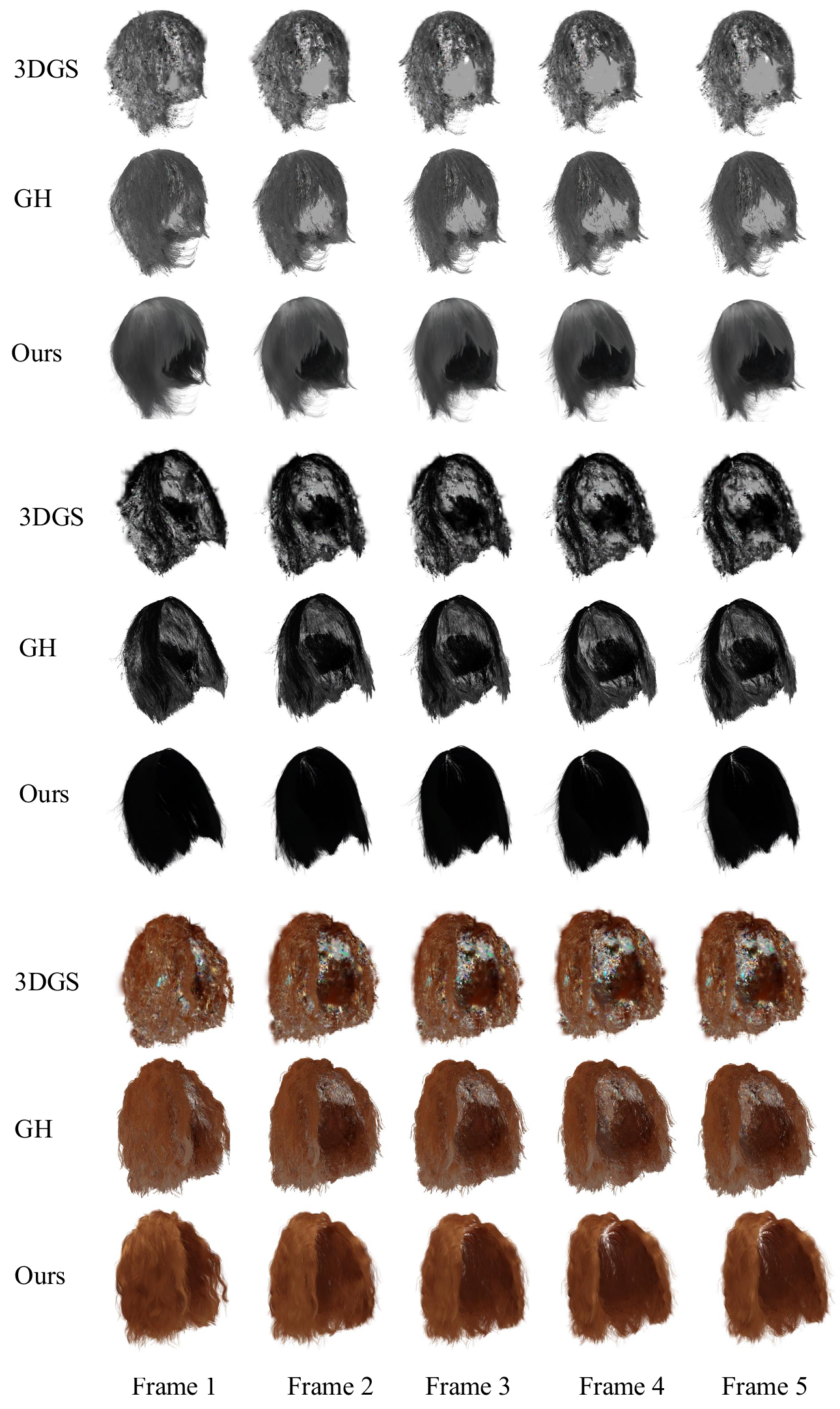}
\caption{\textbf{Comparison results for dynamic hair rendering}}
\label{fig:compare1}
\end{figure*}

\begin{figure*}[htb]
\centering
\includegraphics[width=0.9\textwidth]{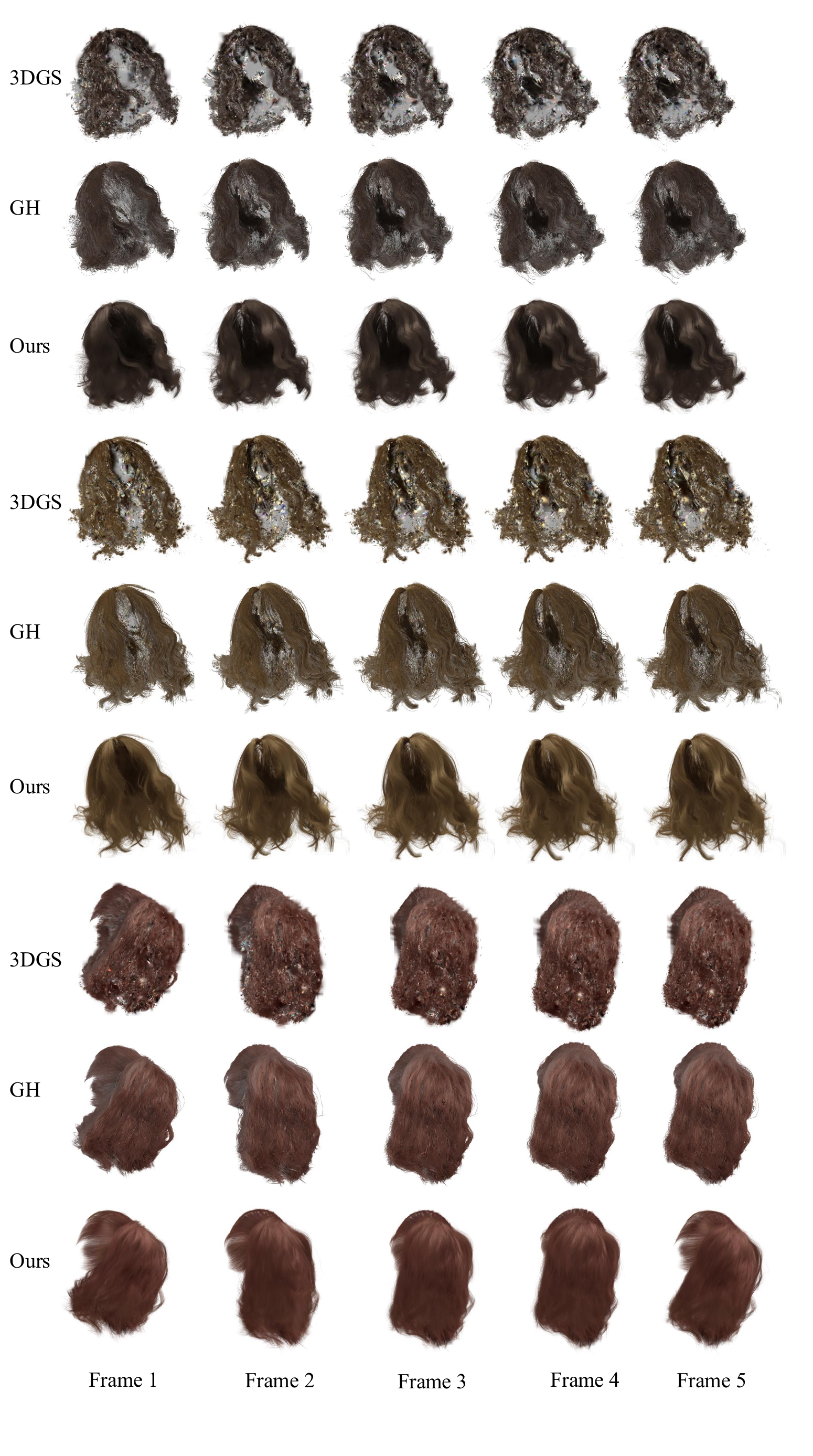}
\caption{\textbf{Comparison results for dynamic hair rendering}}
\label{fig:compare2}
\end{figure*}

\subsection{Dynamic Hair Novel-View Synthesis.}
\textbf{Baseline settings.} We compare our DGH model with two baseline methods: 3DGS~\cite{kerbl20233d} and Gaussian Haircut~\cite{zakharov2024human}. We train both baselines using their publicly released code on our canonical synthetic hair dataset, which includes 48 camera views covering the full hair.
For 3DGS, we initialize Gaussian primitives by attaching them to the center of each hair segment. To maintain consistency with our dynamic sequences, we fix the number of Gaussians and disable density growth. We fix the opacity and positions of Gaussians, compute their orientation using hair tangents, and optimize three degrees of freedom of scale, color (represented using spherical harmonics). We use the Adam optimizer with a learning rate of 0.00025 for SH color features and 0.0005 for scale optimization.
For Gaussian Haircut, we similarly optimize both color and scale, and the scale is constrained to be proportional to the length of each line segment to better capture hair geometry.
Once each baseline model is trained on the static canonical hairstyle, we propagate the learned Gaussian parameters to dynamic sequences using our hair tracking model, which provides per-frame hair geometry. 

\textbf{Comparison and additional results.} We present further results on dynamic hair novel-view synthesis in Fig.~\ref{fig:compare1} and Fig.~\ref{fig:compare2}. To better evaluate appearance quality, we use ground-truth hair tracking and compare our appearance model against other Gaussian-based hair representation methods. Results are shown across multiple dynamic sequences and diverse hairstyles.
\subsection{Multi-Layer Gaussian Avatars with Dynamic Hair.}
\begin{figure*}[t]
\centering
\includegraphics[width=\textwidth]{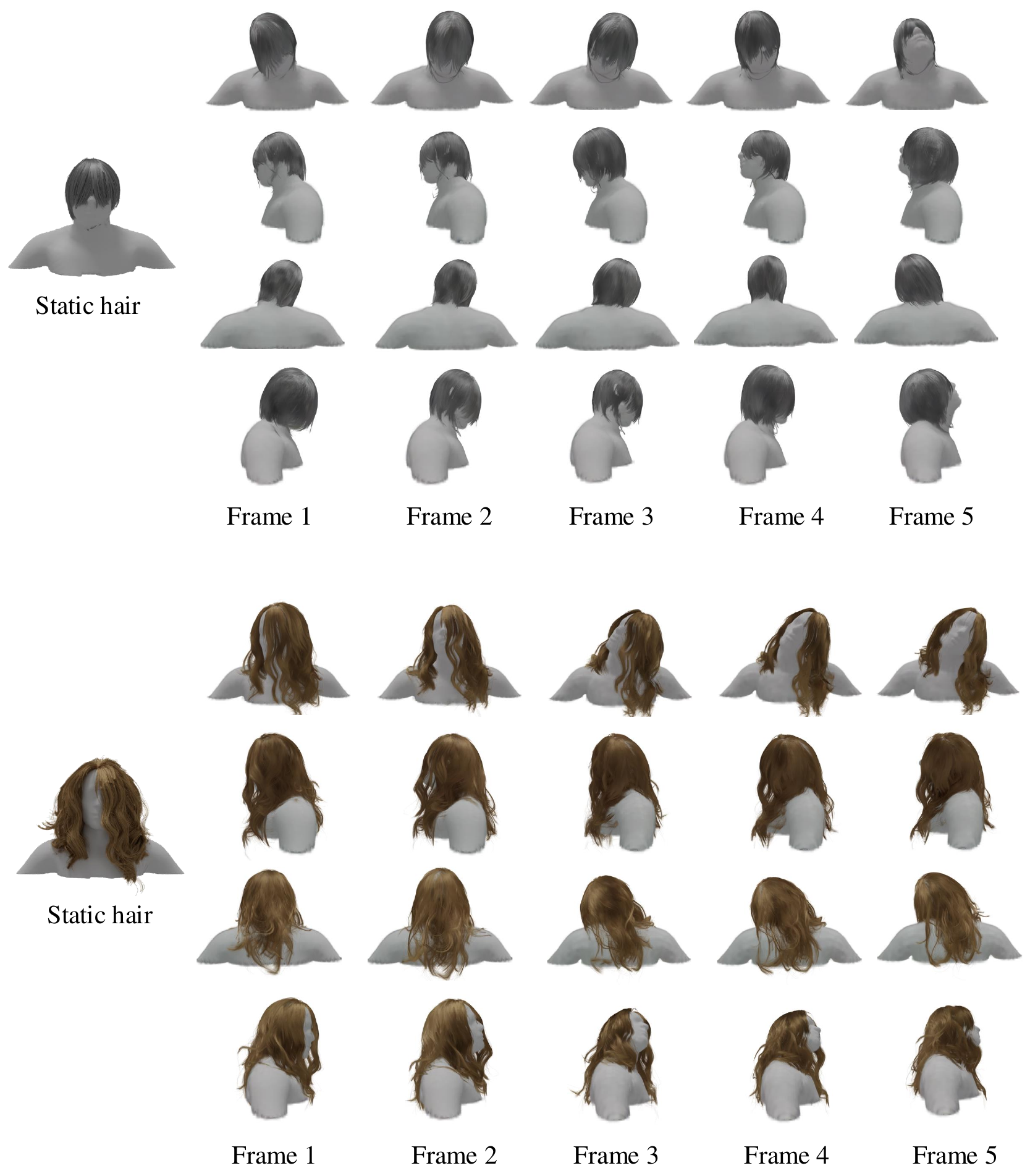}
\caption{Our dynamic Gaussian hair appearance model enables novel view synthesis from static hair using our hair tracking model. Note: Static hair lacks gravity, while the dynamic model implicitly learns gravity from the dataset distribution.}
\label{fig:ours_render1}
\end{figure*}

\begin{figure*}[t]
\centering
\includegraphics[width=\textwidth]{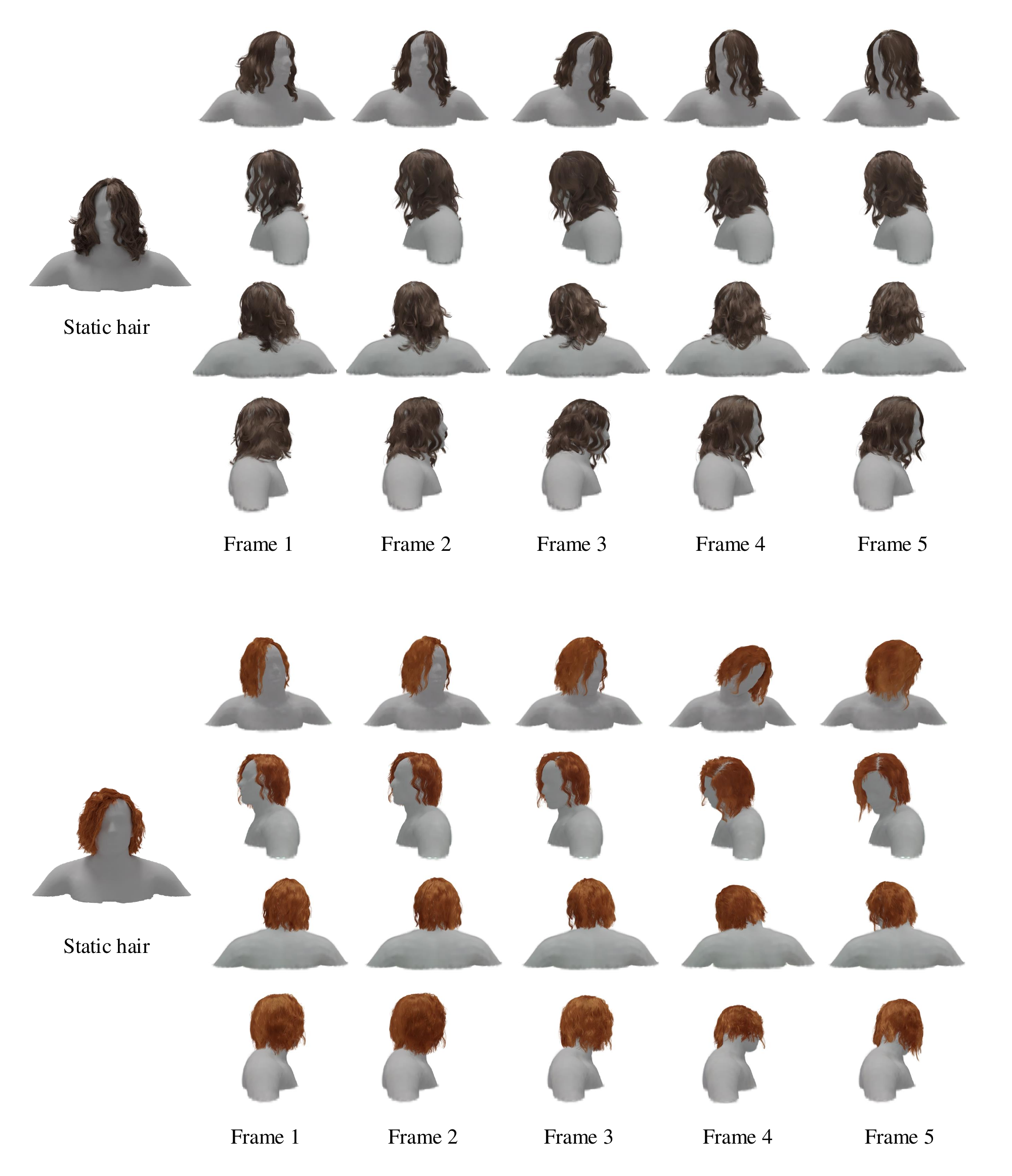}
\caption{Novel view synthesis results from static hair using our hair tracking model across various hair grooms.}
\label{fig:ours_render2}
\end{figure*}

\begin{figure*}[t]
\centering
\includegraphics[width=\textwidth]{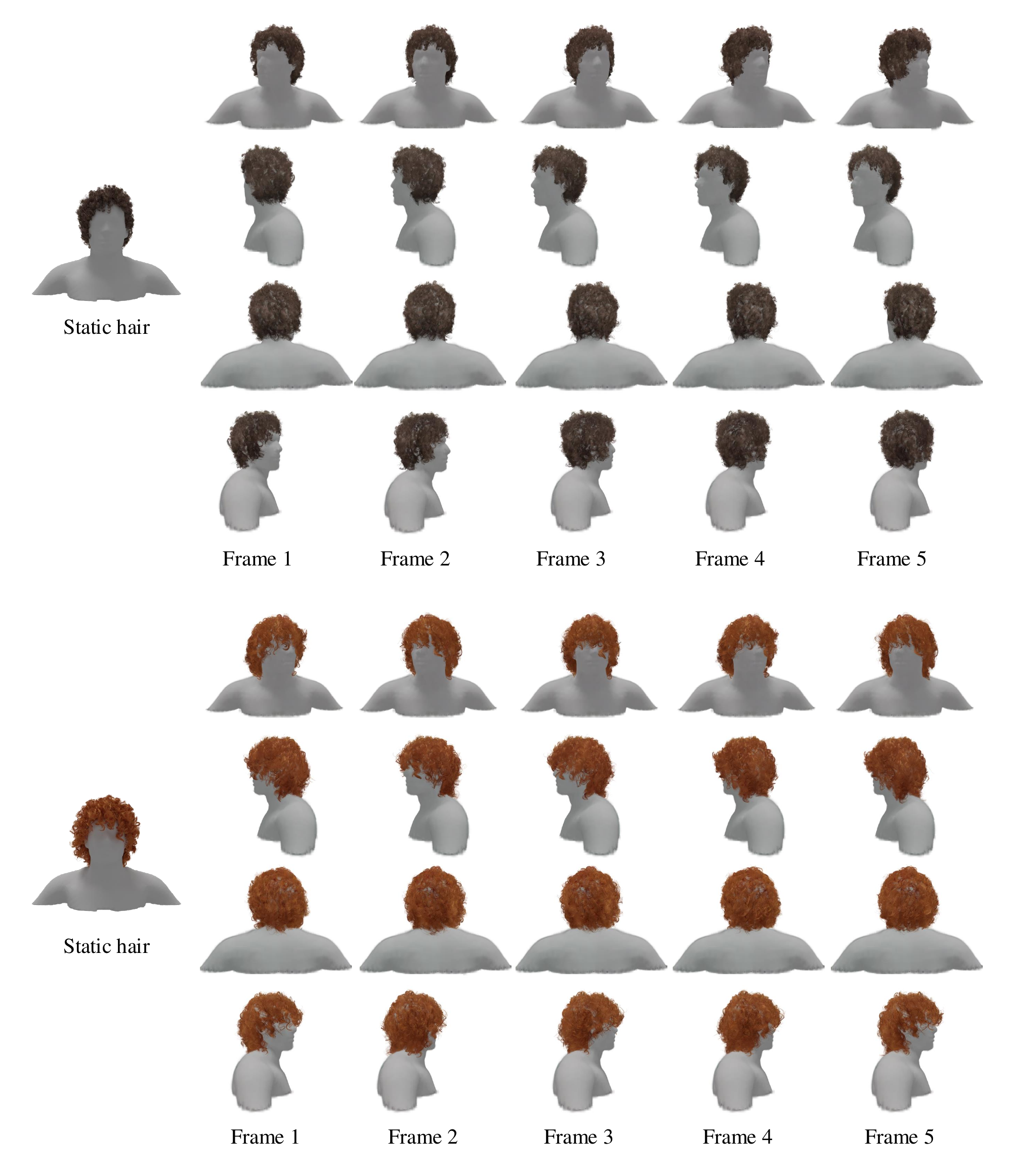}
\caption{Given our hair tracking model and body pose, our framework enables reanimation and novel-view rendering with hair dynamics, including curly hair.}
\label{fig:ours_render3}
\end{figure*}

\begin{figure*}[t]
\centering
\includegraphics[width=\textwidth]{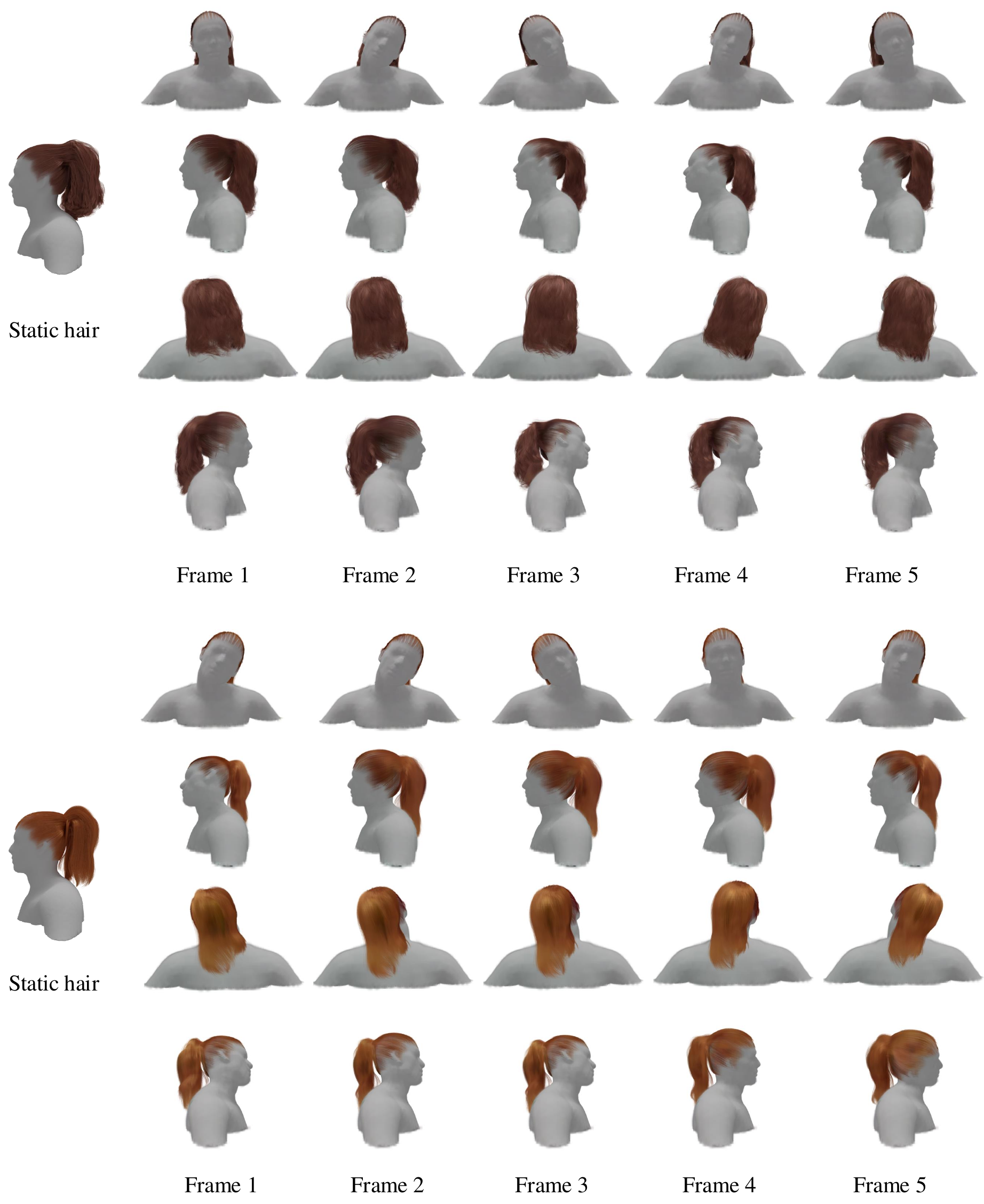}
\caption{Given our hair tracking model and body pose, our framework enables reanimation and novel-view rendering with hair dynamics, including ponytail hairstyles.}
\label{fig:ours_render4}
\end{figure*}

Our dynamic hair Gaussian representation can be seamlessly integrated with pre-trained Gaussian avatar models for avatar re-animation and novel-view rendering, as shown in Fig.~\ref{fig:ours_render1}, \ref{fig:ours_render2}, \ref{fig:ours_render3}, \ref{fig:ours_render4} the render image resolution is 1024 $\times$ 1024. Our appearance model supports only hair rendering, while the body is rendered using a separate pre-trained Gaussian model. For visualization, both hair and body Gaussians are rendered together. \textbf{Note that the static hair lacks gravity effects.} Using our hair tracking model, we represent hair strands as connected cylindrical Gaussians and achieve high-quality novel view synthesis through our motion-dependent appearance model. Unlike static hair, our dynamic hair model implicitly learns gravity effects from the dataset distribution, ensuring realistic rendering across dynamic hair motion sequences. 

In Fig.~\ref{fig:real_human}, we further demonstrate how \dgh can be seamlessly integrated with a multi-layer Gaussian avatar for realistic avatar re-animation. Since our \dgh model primarily takes head motion as input, the resulting motion focuses on head-driven hair dynamics without incorporating body motion effects. Inspired by recent advancements in multi-layer Gaussian avatar modeling~\cite{li2025simavatar}, realistic avatar editing and re-animation can be achieved by introducing additional Gaussian layers to represent various components such as hair and clothing. To merge an additional hair layer with the body avatar, our framework begins with a pre-trained head GS avatar re-animated using head motion inputs. The \dgh module predicts dynamic Gaussian hair deformations directly from a half-body GS representation or a dense point cloud, which are then merged with the head GS to produce realistic, data-driven hair motion. 

In contrast, physics-based methods re-animate a GS avatar by converting it into an explicit mesh, simulating hair motion with a physics engine, rendering it through a 3D pipeline, and re-converting the result into a 3DGS format before merging. Unlike such mesh-based pipelines that require multiple conversion and rendering stages, our DGH framework operates entirely in the Gaussian domain. By representing both hair and the upper body as volumetric Gaussians, it enables mesh-free deformation prediction, ensuring compatibility with Gaussian-based avatars and seamless integration into learning-based re-animation pipelines without rigging or simulation overhead.

\section{Limitation and Future Work} 
DGH currently handles hair-upper-body collisions. Environmental constraints and relighting of DGH (via albedo inference) could be handled in future work. Due to the challenges of accurately tracking real hair dynamics, precise deformation and per-timestep positions of hair strands are difficult to record. Our model is currently trained and evaluated on synthetic data, focusing on generalizing to head motions. In future work, we aim to create a large dynamic hair dataset based on static 3D hair datasets~\cite{digitalsalon, he2024perm}, leveraging a learned deformed hair prior. This identity-independent model could robustly infer diverse hairstyles and perform well on real-world data. 

In the second stage, a lightweight MLP combined with a differentiable rasterizer achieves high-fidelity novel view rendering, balancing efficiency and quality. While fast for dense Gaussian primatives, the current version is not real-time for dense hair (100K-150K hair strands). Future work could focus on optimizing the MLP or leveraging caching methods~\cite{muller2021real, garbin2021fastnerf}, such as viewpoint-aware or temporal caching, to achieve real-time performance.

For multi-layer Gaussian avatars with dynamic hair applications, our model currently supports integration with head avatars driven solely by head motion. Full-body effects such as jumping and squirming under gravity are not yet modeled. In future work, we plan to incorporate explicit material conditioning and full-body motion inputs to improve generalization across diverse hairstyles, particularly with access to a more comprehensive 3D hairstyle dataset.

\section{Broader Impacts} 
As a positive impact, our work enhances the realism of Gaussian-based digital humans, enabling more expressive virtual interactions in telepresence, AR/VR, and virtual production. As a negative impact, increased photorealism may also raise ethical concerns, including potential misuse for identity manipulation or deceptive content generation.